\documentclass[10pt,twocolumn,letterpaper]{article}

\usepackage[pagenumbers]{cvpr} 

\usepackage[table,dvipsnames]{xcolor}

\usepackage{lipsum}
\usepackage{subcaption}
\usepackage{tikz}
\usepackage{booktabs}
\usepackage{multirow}
\usepackage{float}
\usepackage{dsfont}
\usepackage{algorithm}
\usepackage[commentColor=black,noEnd=false]{algpseudocodex}
\usepackage{pifont}
\newcommand{\cmark}{\text{\ding{51}}}
\newcommand{\xmark}{\text{\ding{55}}}

\newcommand{\squeezeup}{\vspace{-2.5mm}}

\definecolor{figred}{HTML}{B85450}
\definecolor{figgreen}{HTML}{82B366}
\definecolor{figblue}{HTML}{6C8EBF}
\definecolor{figyellow}{HTML}{D6B656}
\definecolor{figpurple}{HTML}{785C85}

\usepackage[accsupp]{axessibility} 

\definecolor{cvprblue}{rgb}{0.21,0.49,0.74}
\usepackage[pagebackref,breaklinks,colorlinks,citecolor=cvprblue]{hyperref}

\def\paperID{9477} 
\def\confName{CVPR}
\def\confYear{2024}
    
\title{Continual Segmentation with Disentangled Objectness Learning\\and Class Recognition}

\author{Yizheng Gong$^{1,2}$ \quad Siyue Yu$^{1}$ \quad Xiaoyang Wang$^{1,2,3}$ \quad Jimin Xiao$^{1,}\thanks{Corresponding author.}$ \vspace{0.3em} \\
{\normalsize $^1$Xi'an Jiaotong-Liverpool University} \quad
{\normalsize $^2$University of Liverpool} \quad
{\normalsize $^3$Metavisioncn} \\
{\footnotesize \texttt{y.gong21@liverpool.ac.uk, siyue.yu@xjtlu.edu.cn, wangxy@liverpool.ac.uk, jimin.xiao@xjtlu.edu.cn}}
}

\begin{document}
\maketitle
\begin{abstract}
Most continual segmentation methods tackle the problem as a per-pixel classification task. However, such a paradigm is very challenging, and we find query-based segmenters with built-in objectness have inherent advantages compared with per-pixel ones, as objectness has strong transfer ability and forgetting resistance. Based on these findings, we propose CoMasTRe by disentangling continual segmentation into two stages: forgetting-resistant continual objectness learning and well-researched continual classification. CoMasTRe uses a two-stage segmenter learning class-agnostic mask proposals at the first stage and leaving recognition to the second stage. During continual learning, a simple but effective distillation is adopted to strengthen objectness. To further mitigate the forgetting of old classes, we design a multi-label class distillation strategy suited for segmentation. We assess the effectiveness of CoMasTRe on PASCAL VOC and ADE20K. Extensive experiments show that our method outperforms per-pixel and query-based methods on both datasets. Code will be available at \url{https://github.com/jordangong/CoMasTRe}.
\end{abstract}    
\section{Introduction}
\label{sec:intro}

Neural nets have dominated computer vision, from object recognition~\cite{heDeepResidualLearning2016}~to fine-grained classification~\cite{Liu2023ProgressiveSM,9656684}, segmentation~\cite{chenRethinkingAtrousConvolution2017,Wang2023HuntingSD}, and detection~\cite{zhangDINODETRImproved2023}. However, the success is only sound for single-shot learning, \ie, training once to perform well with fully annotated datasets without considering the later learning process. \textit{Continual learning} aims to mimic human learning by gradually obtaining knowledge in a sequential fashion~\cite{Fagot2006EvidenceFL,Grutzendler2002LongtermDS}. Particularly, continual learning shines in dense prediction tasks, such as semantic segmentation~\cite{cermelliModelingBackgroundIncremental2020,douillardPLOPLearningForgetting2021,cermelliCoMFormerContinualLearning2023}, since the annotations are laborious to obtain, and sometimes the learned samples are inaccessible, such as the nature of stream data in autonomous driving~\cite{neuholdMapillaryVistasDataset2017}~or patient privacy in medical imaging~\cite{ozdemirExtendingPretrainedSegmentation2019}.

\begin{figure}
  \begin{subfigure}[t]{\columnwidth}
    \includegraphics[width=\linewidth]{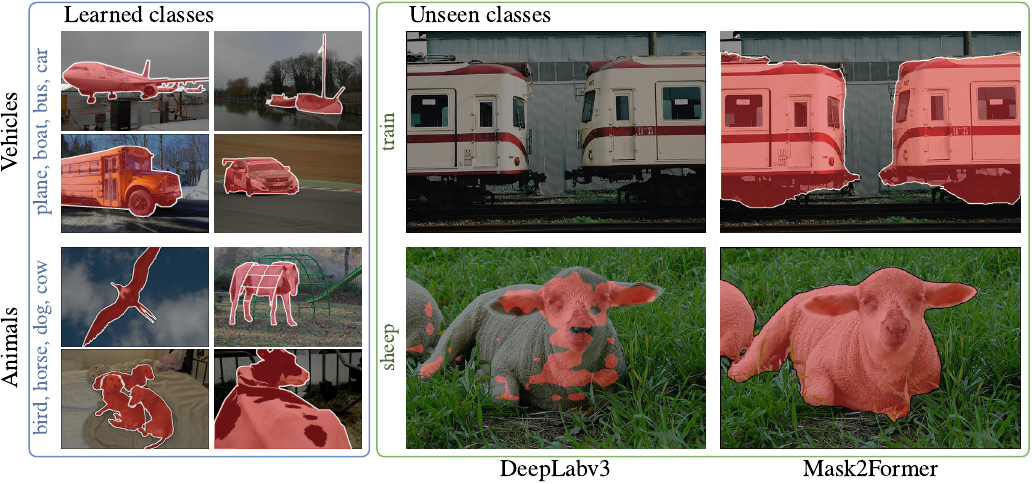}
    \caption[obj-general]{Mask proposal transferred to unseen classes.}
    \label{fig:obj-general}
  \end{subfigure}
  \begin{subfigure}[t]{\columnwidth}
    \includegraphics[width=\linewidth]{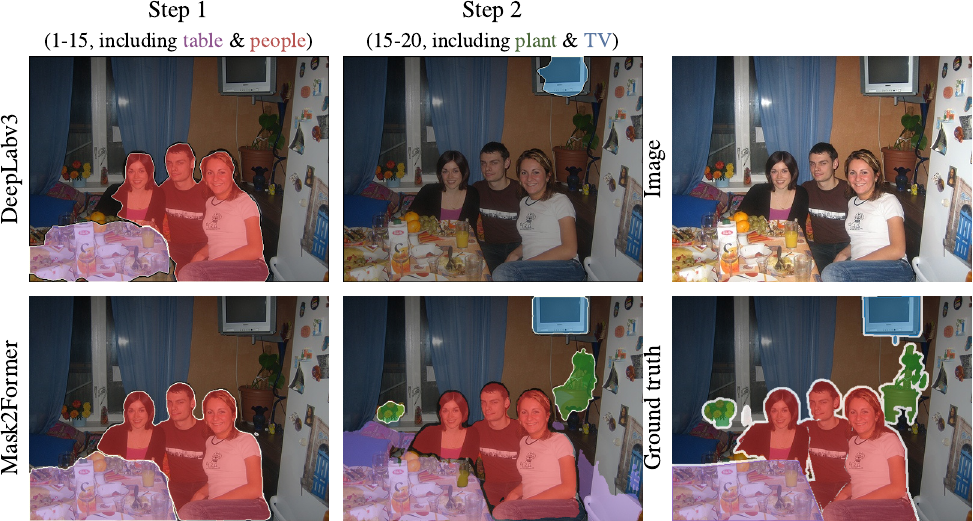}
    \caption[obj-forget]{Mask proposals after finetuning on new classes.}
    \label{fig:obj-forget}
  \end{subfigure}
  \vspace{-2mm}
  \caption[obj-props]{\textbf{Hidden properties inside query-based segmenters.} Objectness in query-based methods helps generalize mask proposals on unseen classes similar to learned classes (top). Additionally, because of the transfer ability of objectness, query-based methods are resistant to catastrophic forgetting of mask proposals (bottom).}
  \squeezeup
  \squeezeup
  \label{fig:obj-props}
\end{figure}

However, machines tend to forget the early tasks if we train them to learn the new ones, also termed as \textit{catastrophic forgetting}~\cite{frenchCatastrophicForgettingConnectionist1999}, which is the critical obstacle toward continual learning. Moreover, dense prediction tasks tackled as per-pixel classification or regression by analyzing patterns around each pixel are inherently difficult, which makes the continual learning of these tasks even more challenging. Could we mitigate forgetting in continual segmentation in a more feasible way? The answer lies in a new paradigm: \textit{mask classification}, streamlining continual segmentation as learning without forgetting of class-agnostic binary masks and class recognition, via query-based segmenters~\cite{chengPerPixelClassificationNot2021,wangMaXDeepLabEndtoEndPanoptic2021,chengMaskedAttentionMaskTransformer2022,yuKmeansMaskTransformer2022}. Different from conventional per-pixel-based segmenters~\cite{chenRethinkingAtrousConvolution2017,strudelSegmenterTransformerSemantic2021,xieSegFormerSimpleEfficient2021}, from our observation, query-based segmenters have strong objectness (the ability to find objects in images) built-in, which is beneficial to continual segmentation for two reasons. Firstly, as the background is not silenced when learning to propose masks, the objectness can be transferred to unseen classes. As illustrated in \cref{fig:obj-general}, compared to DeepLabv3~\cite{chenRethinkingAtrousConvolution2017}, a per-pixel segmentation baseline, query-based Mask2Former~\cite{chengMaskedAttentionMaskTransformer2022}~proposes substantially better masks on unseen classes if the model has learned similar classes before, \eg, learning vehicles, such as planes, boats, buses, and cars, facilitates the later learning of trains. Secondly, we observe that the objectness alleviates the forgetting of old class mask proposals (see \cref{fig:obj-forget}), as the model can still propose comparable masks on old classes, even finetuning on new classes.

Based on these findings, we propose \textbf{Co}ntinual Learning with \textbf{Mas}k-\textbf{T}hen-\textbf{Re}cognize \textbf{TR}ansformer decoder, termed as \textbf{CoMasTRe}, for continual segmentation. CoMasTRe disentangles the segmentation task to objectness learning and class recognition by learning to propose class-agnostic masks at the first stage and leaving recognition to the second stage. In continual settings, since CoMasTRe inherits the objectness of query-based methods, the learning of new mask proposals is considerably eased. This design also simplifies continual segmentation to forgetting-resistant continual objectness learning and a continual classification task. Specifically, CoMasTRe mitigates forgetting in two folds: (i) a simple but effective objectness distillation to reinforce old class objectness during long learning processes, and (ii) a multi-label class distillation strategy with task-specific classifiers to alleviate old class forgetting. For evaluation, we conduct extensive experiments on PASCAL VOC 2012 and ADE20K, showcasing that CoMasTRe reaches a new state-of-the-art on both datasets. In particular, we boost incremented class performance compared with previous best methods on VOC. Additionally, CoMasTRe significantly outperforms prior arts in all classes on ADE20K.

To sum up, the contributions of this paper are as follows:

\begin{itemize}[leftmargin=*,labelindent=13pt]
  \item To leverage the properties of objectness, we propose \textbf{Co}ntinual learning with \textbf{Mas}k-\textbf{T}hen-\textbf{Re}cognize \textbf{TR}ansformer decoder, termed as CoMasTRe, to simply continual segmentation by decoupling the task into objectness learning and class recognition.
  \item To tackle the forgetting issue, we strengthen the objectness using a simple but effective distillation strategy in the first stage and preserve class knowledge in the second stage with task-specific classifiers and multi-label class distillation tailored to segmentation.
  \item Through extensive benchmarking on two datasets, our method outperforms prior query-based methods by up to $2.1\%$ on ADE20K and surpasses per-pixel methods on new classes by up to $32.16\%$ on PASCAL VOC.
\end{itemize}
\section{Related Work}
\label{sec:related-work}

\textbf{Query-based Image Segmentation.}
Query-based segmenters~\cite{chengPerPixelClassificationNot2021,chengMaskedAttentionMaskTransformer2022,yuKmeansMaskTransformer2022,jainOneFormerOneTransformer2023}~have unified image segmentation via \textit{mask classification} and solve semantic, instance, and panoptic segmentation using the same framework.
However, the \textit{mask classification} paradigm was originated from query-based detectors, such as DETR~\cite{carionEndtoEndObjectDetection2020a} and its variants~\cite{zhuDeformableDETRDeformable2021,liDNDETRAccelerateDETR2022,zhangDINODETRImproved2023},
where we train the model to predict proposals and class labels with learnable queries. MaskFormer~\cite{chengPerPixelClassificationNot2021}~was the first to introduce \textit{mask classification} in semantic segmentation, which was dominated by per-pixel classification paradigm via convolutional nets~\cite{chenRethinkingAtrousConvolution2017,xiaoUnifiedPerceptualParsing2018,wangAxialDeepLabStandAloneAxialAttention2020}~or Transformers~\cite{strudelSegmenterTransformerSemantic2021,xieSegFormerSimpleEfficient2021}.
Following MaskFormer,
Mask2Former~\cite{chengMaskedAttentionMaskTransformer2022}~introduced multiple improvements, such as masked attention for mask refinement and using multi-scale features for better small object performance. To speed up the convergence, $k$MaX-DeepLab~\cite{yuKmeansMaskTransformer2022} generated queries from clustered image features. Recently, OneFormer~\cite{jainOneFormerOneTransformer2023} provided a generalist solution by joint training of all segmentation tasks. However, all of these segmenters learn to predict to mask and class simultaneously and are not suitable for decoupled continual segmentation.

\noindent\textbf{Continual Segmentation.}
Learning new knowledge by fintuning the model is a native way, but resulting \textit{catastrophic forgetting} of old knowledge~\cite{frenchCatastrophicForgettingConnectionist1999, mccloskeyCatastrophicInterferenceConnectionist1989, robinsCatastrophicForgettingRehearsal1995}. Continual learning aims to alleviate the forgetting issue by balancing the learning of new knowledge (\textit{plasticity}) and the preserving of the old one (\textit{stability}).
While the research focus of continual learning is on classification~\cite{kirkpatrickOvercomingCatastrophicForgetting2017,liLearningForgetting2018,riemerLearningLearnForgetting2019,buzzegaDarkExperienceGeneral2020,douillardPODNetPooledOutputs2020,chaCo2LContrastiveContinual2021,wangLearningPromptContinual2022,smithCODAPromptCOntinualDecomposed2023,Zhang_2023_ICCV}, continual segmentation started to thrive recently~\cite{cermelliModelingBackgroundIncremental2020,douillardPLOPLearningForgetting2021,michieliContinualSemanticSegmentation2021,chaSSULSemanticSegmentation2021,phanClassSimilarityWeighted2022,zhangRepresentationCompensationNetworks2022,zhangMiningUnseenClasses2022,cermelliCoMFormerContinualLearning2023,shangIncrementerTransformerClassIncremental2023,xiaoEndpointsWeightFusion2023,zhuContinualSemanticSegmentation2023,zhangCoinSegContrastInter2023}.
As pointed out in MiB~\cite{cermelliModelingBackgroundIncremental2020}, the background shift in continual segmentation aggravates the forgetting issue. Cermelli \etal~\cite{cermelliModelingBackgroundIncremental2020}~proposed unbiased distillation and classification to mitigate the background shift. PLOP~\cite{douillardPLOPLearningForgetting2021}~circumvented the background shift issue by feature distillation and pseudo-labeling. Based on PLOP, REMINDER~\cite{phanClassSimilarityWeighted2022} introduced class-weighted knowledge distillation to discriminate the new and old classes. With the same motivation, Incrementer~\cite{shangIncrementerTransformerClassIncremental2023}~designed a class deconfusion strategy and achieved competitive performance via a Transformer-based segmenter. Recently, RL-Replay~\cite{zhuContinualSemanticSegmentation2023}~set a strong replay-based baseline via reinforcement learning. Some works~\cite{cermelliIncrementalLearningSemantic2022,hsiehClassincrementalContinualLearning2023}~started to focus on more realistic scenarios by integrating weak supervision~\cite{Zhao2024SFCSF,Zhang2023CredibleDL}~with continual learning. The methods above follow the \textit{per-pixel} paradigm. Alhough CoMFormer~\cite{cermelliCoMFormerContinualLearning2023} was the first to present \textit{mask classification} in continual segmentation, it applied a standard framework with distillation and pseudo-labeling and failed to leverage the benefit of objectness. In contrast, CoMasTRe exploits objectness and decouples continual segmentation into forgetting-resistant objectness learning and class recognition.

\noindent\textbf{Continual Dynamic Networks.}
In addition to distillation methods appearing in continual segmentation, dynamic networks with parameter expansion~\cite{yanDynamicallyExpandableRepresentation2021,wangFOSTERFeatureBoosting2022,zhouModel603Exemplars2023,douillardDyToxTransformersContinual2022,wangLearningPromptContinual2022,wangDualPromptComplementaryPrompting2022,smithCODAPromptCOntinualDecomposed2023,smithCODAPromptCOntinualDecomposed2023,razdaibiedinaProgressivePromptsContinual2023}~also play a key role in continual learning. DER~\cite{yanDynamicallyExpandableRepresentation2021}~is an early work on the dynamic structure by duplicating the backbone for each task to mitigate the forgetting issue. However, DER suffers from the explosion of parameters after a long learning process. FOSTER~\cite{wangFOSTERFeatureBoosting2022} relieved the problem through a model compression stage. MEMO~\cite{zhouModel603Exemplars2023} went further by decoupling the network and only expanding specialized layers. Recently, more works~\cite{douillardDyToxTransformersContinual2022,wangLearningPromptContinual2022,wangDualPromptComplementaryPrompting2022,smithCODAPromptCOntinualDecomposed2023,smithCODAPromptCOntinualDecomposed2023,razdaibiedinaProgressivePromptsContinual2023}~started to leverage the dynamic nature of Transformers. DyTox~\cite{douillardDyToxTransformersContinual2022} proposed a dynamic token expansion mechanism by learning a task-specialized prompt at each step. The concurrent work, L2P~\cite{wangLearningPromptContinual2022}, started a new fashion of continual learning: learning to prompt pretrained models and extracting task-specialized features. Thanks to our Transformer-based decoder, we adopt task queries and task-specific classifiers in our class decoder to further alleviate forgetting.

\section{Method}
\label{sec:method}

\subsection{Problem Definition}
\label{subsec:prob-def}

Image segmentation has been unified as a \textit{mask classification} problem, \ie, to propose class-agnostic masks and predict the corresponding class labels simultaneously. Formally, we define the dataset $\mathcal{D}$ containing pairs of image and target, and each sample pair contains an image $x \in \mathbb{R}^{C \times H \times W}$ and its target $y$, where $C$, $H$ and $W$ denote channel number, height and width of the image, respectively. Each target $y$ is composed by $M$ ground truth (GT) binary masks and class labels, denoted $y = \{(m_i^\mathrm{gt}, c_i^\mathrm{gt}) \mid m_i^\mathrm{gt} \in \{0, 1\}^{H \times W}, c_i^\mathrm{gt} \in \mathcal{C}\}_{i=1}^{M}$, where $\mathcal{C}$ is the class label space. Unlike per-pixel classification methods, background class is excluded from annotation. 

In terms of continual segmentation, the model sequentially learns to predict the masks of new classes but not to forget old ones. Formally, we divide the learning process to $T$ tasks, and at each step $t = 1, 2, \dots, T$ we train the model to predict a unique set of classes $\mathcal{C}^t$, where $\bigcap_{t=i}^{T} \mathcal{C}^{t} = \varnothing$ and $\bigcup_{t=i}^{T} \mathcal{C}^{t} = \mathcal{C}$. Only the part of the annotations containing new classes $\mathcal{C}^t$ are provided at step $t$, while the model should be able to predict all learned classes $\mathcal{C}^{1:t}$.
 
\subsection{CoMasTRe Architecture}
\label{subsec:comastre-arch}


\begin{figure}
    \includegraphics[width=\columnwidth]{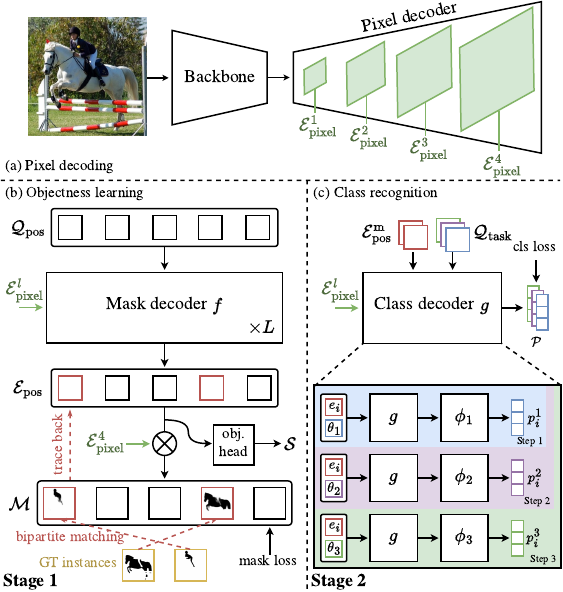}
    \vspace{-5mm}
    \caption[obj]{\textbf{CoMasTRe Architecture.} $\bigotimes$ denotes the dot product between positional embeddings $\mathcal{E}_\mathrm{pos}$ and pixel embeddings $\mathcal{E}^4_\mathrm{pixel}$.
    CoMasTRe uses a two-stage image segmenter including three components: (a) a backbone and a pixel decoder producing pixel embedding, (b) a mask decoder $f$ with learnable positional queries $\mathcal{Q}_\mathrm{pos}$ for objectness learning, and (c) a class decoder $g$ with a set of task queries $\mathcal{Q}_\mathrm{task}$ for class recognition.
    }
    \vspace{-4mm}
    \label{fig:obj}
\end{figure}

Thanks to the properties of objectness (see \cref{fig:obj-props}) and widely studied continual classification, we argue that solving the continual segmentation problem with \textit{mask classification} is inherently advantageous. Therefore, we propose CoMasTRe, which adopts Mask2Former~\cite{chengMaskedAttentionMaskTransformer2022}~meta-architecture, keeping the backbone and the pixel decoder while adding two newly designed Transformer decoders to help disentangle objectness learning (stage 1) and class recognition (stage 2), as shown in \cref{fig:obj}. The overall training process of our CoMasTRe can be summarized as:
\begin{enumerate}
    \item When an image comes to CoMasTRe, the backbone, and the pixel decoder first encode and then decode the image to pixel embeddings. We take out the pixel embeddings from all the layers of the pixel decoder for the following process as $\{\mathcal{E}_\mathrm{pixel}^l\}^4_{l=1}$, where $l$ denotes different layers and there are 4 layers in total  (\cref{fig:obj} (a)). 

    \item Next, we randomly initialize learnable positional queries $\mathcal{Q}_\mathrm{pos}$ and input them to the objectness learning stage (stage 1), to extract positional embeddings $\mathcal{E}_\mathrm{pos}$ through the mask decoder $f$ for predicting class-agnostic mask proposals $\mathcal{M}$ and objectness scores $\mathcal{S}$. The training involves bipartite matching with ground truth (\cref{fig:obj} (b)).
    
    \item After that, matched positional embeddings $\mathcal{E}_\mathrm{pos}^\mathrm{m}$ from stage 1 with pixel embeddings are input to class decoder $g$ for recognition (stage 2). To reduce task interference during continual learning, task queries $\mathcal{Q}_\mathrm{task}$ and task-specific classifiers $\phi_1, \dots, \phi_t$ are learned to specialize the class knowledge of each task. (\cref{fig:obj} (c)). 

    \item Finally, we obtain segmentation results by combining the mask proposals and objectness scores from stage 1 with the class prediction from stage 2.
\end{enumerate}

The details of the two stages will be illustrated in \cref{para:stage-1} and \cref{para:stage-2}.

\subsubsection{Stage 1: Objectness Learning}
\label{para:stage-1}
As illustrated in \cref{fig:obj} (b), our objectness learning mainly relies on a mask decoder. The mask decoder $f$ is composed of $L$ blocks of Transformer layers~\cite{vaswaniAttentionAllYou2017}, taking $N$ learnable positional queries $\mathcal{Q}_\mathrm{pos} = \{q_1, \dots, q_N\} \in \mathbb{R}^{N \times d}$ and intermediate pixel embeddings $\{\mathcal{E}^l_\mathrm{pixel}\}_{l=1}^3$ as input. Then, it outputs the positional embeddings $\mathcal{E}_\mathrm{pos} = \{e_1, \dots, e_N\} \in \mathbb{R}^{N \times d}$ for mask proposals, objectness scores, and stage 2 recognition. The mask proposals $\mathcal{M} = \{m_1, \dots, m_N\} \in [0, 1]^{N \times H \times W}$ is computed as $\mathcal{M} = \mathrm{sigmoid}(\mathrm{Upsample}(\mathrm{MLP}(\mathcal{E}_\mathrm{pos}) \cdot \mathcal{E}_\mathrm{pixel}^4))$, where the $\mathrm{MLP}(\cdot)$ acts as a non-linear transformation function and $\mathrm{Upsample}(\cdot)$ interpolates the logits to image size. Note here each positional embedding produces a mask proposal. At the same time, $\mathcal{E}_\mathrm{pos}$ is input to our objectness head, in which a binary classifier is designed to output the objectness score $\mathcal{S} = \{s_1, \dots, s_N\} \in [0, 1]^N$ to indicate whether the mask proposals containing objects or not. During training, we first perform bipartite matching considering the cost between $N$ mask proposals $\{m_i\}_{i=1}^N$ and $M$ ground truth masks $\{m_j^\mathrm{gt}\}_{j=1}^M$. Here, we assume $N\gg M$ and pad the ground truth with ``no object'' $\varnothing$ to ensure a one-to-one matching between predictions and ground truth. 
After solving the matching via the Hungarian algorithm~\cite{kuhnHungarianMethod}, we get the optimal permutation of $N$ elements, denoted as $\sigma$. Then, we optimize this stage using ground truth masks:
\begin{equation}
\begin{aligned}
    \mathcal{L}_\mathrm{obj} = \sum_{j=1}^N \big[&-\mathds{1}_{m_j^\mathrm{gt} \neq \varnothing}{\log{s_{\sigma_j}}} - \mathds{1}_{m_j^\mathrm{gt} = \varnothing}{\log{(1-s_{\sigma_j})}} \\
    &+\mathds{1}_{m_j^\mathrm{gt} \neq \varnothing}\mathcal{L}_\mathrm{mask}(m_{\sigma_j}, m_j^\mathrm{gt})\big],
\end{aligned}
\end{equation}
where $\mathcal{L}_\mathrm{mask}$ is the sum of binary cross-entropy loss and Dice loss~\cite{sudreGeneralisedDiceOverlap2017}.
Based on the matching results, $M$ matched positional embeddings $\mathcal{E}^\mathrm{m}_\mathrm{pos} =
\{e_j \mid e_j \in \mathcal{E}_\mathrm{pos}, m_j^\mathrm{gt} \neq \varnothing\}$ are traced back and fed forward to stage 2 class decoder. Here, we supervise the objectness scores by encouraging the matched ones and suppressing the unmatched ones, otherwise, the model tends to produce high objectness scores for all the proposals. The remaining unmatched embeddings $\mathcal{E}^\mathrm{u}_\mathrm{pos} = \mathcal{E}_\mathrm{pos} \setminus \mathcal{E}^\mathrm{m}_\mathrm{pos}$ are reserved for distillation to alleviate forgetting (see \cref{subsec:comastre-distill}).

\subsubsection{Stage 2: Class Recognition}
\label{para:stage-2}

Our stage 2 aims to recognize the object inside the mask proposals. Thus, we use matched positional embeddings along with pixel embeddings as input for class recognition. Besides, inspired by DyTox~\cite{douillardDyToxTransformersContinual2022}, we integrate task queries with matched positional embeddings as input to task-specific classifiers in the class decoding process, which facilitates continual learning by reducing task interference, resulting in more specialized classifiers.

Since continual segmentation contains $T$ tasks, to illustrate our class recognition stage, we assume the current learning step is $t$. 
We denote task queries $\mathcal{Q}_\mathrm{task} = \{\theta_1, \dots, \theta_t\}$. As shown in \cref{fig:obj} (c), for each positional embedding $e \in \mathcal{E}_\mathrm{pos}^\mathrm{m}$ and each task query $\theta \in \mathcal{Q}_\mathrm{task}$, we take $(e, \theta)$ as a pair and feed it with pixel embedding to the class decoder $g$ (a Transformer block shared across tasks), obtaining the localized task embedding $k$. For each positional embedding, we obtain its task embeddings $\mathcal{E}_\mathrm{task} = \{k_1, \dots, k_t\}$. Then, $t$ task-specific classifiers $\Phi^t = \{\phi_1, \dots, \phi_t\}$ are applied to $\mathcal{E}_\mathrm{task}$, where each classifier $\phi_j \in \Phi^t$ is specialized to predict the classes in this task $\mathcal{C}^j$, \ie, $\phi_j: \mathbb{R}^d \mapsto {\mathcal{C}^j}$. In this way, we obtain the class probability of each proposal as $p = \mathrm{sigmoid}([\phi_1(k_{1}), \dots, \phi_t(k_{t})])$, as each positional embedding corresponds to a mask proposal.
By feeding all the positional embeddings, we get the class probability of all proposals $\mathcal{P} = \{p_i\}_{i=1}^{M}$. We supervise this stage with classification loss using ground truth class labels:
\begin{equation}
\begin{aligned}
    \mathcal{L}_\mathrm{cls} = - \frac{1}{M}\sum_{i=1}^M \sum_{c \in \mathcal{C}^{t}}\big[&\mathds{1}_{c = c_i^\mathrm{gt}}(1 - p_{i,c})^\gamma \log{p_{i,c}} \\
    + &\mathds{1}_{c \neq c_i^\mathrm{gt}}p_{i,c}^\gamma \log{(1 - p_{i,c})}\big],
\end{aligned}
\end{equation}
where $p_{i,c}$ is the predicted probability of class $c$ for the $i$-th proposal, $\gamma$ is a hyperparameter. Here, we use focal loss~\cite{linFocalLossDense2017}~instead of cross-entropy loss for better calibration~\cite{mukhotiCalibratingDeepNeural2020}, which smooths the distillation process during continual learning. The overall segmentation loss is the sum of objectness loss and classification loss, \ie, $\mathcal{L}_\mathrm{seg} = \mathcal{L}_\mathrm{obj} + \mathcal{L}_\mathrm{cls}$. We jointly train two stages by minimizing $\mathcal{L}_\mathrm{seg}$. During inference, we set an objectness threshold $\alpha$ at stage 1 to filter low confident prediction and feed high objectness embedding to stage 2 for mask recognition.


\subsection{Learning without Forgetting with CoMasTRe}
\label{subsec:comastre-distill}

Learning new classes by naively finetuning the model will cause catastrophic forgetting. To alleviate this issue, our CoMasTRe separately considers the distillation of both stages. In this way, predictions on base classes can also be preserved when learning new classes.


\subsubsection{Objectness Distillation}
\label{sec:obj-distill}
\begin{figure}
    \includegraphics[width=\columnwidth]{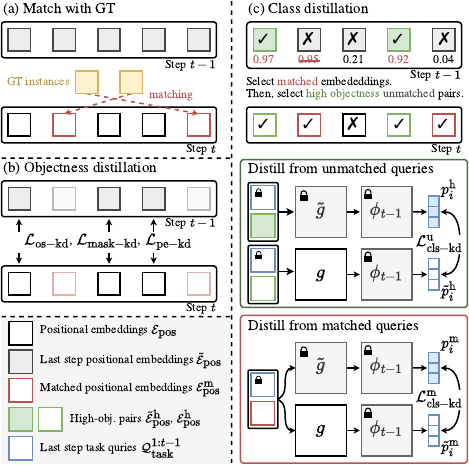}
    \vspace{-5mm}
    \caption[obj-dist]{\textbf{Learning without forgetting with CoMasTRe.} To tackle catastrophic forgetting, CoMasTRe separated the distillation process into two stages, including objectness distillation and class distillation. We perform bipartite matching first, as in (a). Then, we distill the knowledge of objectness for remaining embeddings, as in (b). Finally, we select positional embeddings for the class distillation if they match with ground truth or have high objectness scores at the last step. The class knowledge is distilled from both matched and unmatched positional embeddings with a class decoder and task-specific classifiers, as in (c).}
    \label{fig:obj-dist}
    \vspace{-4mm}
\end{figure}

As shown in \cref{fig:obj-dist} (a), we first train the model with bipartite matching and get matched positional embeddings and unmatched ones. However, the objectness scores of unmatched ones quickly diminish as the ground truth does not contain masks from old classes, resulting in incorrect objectness scores on old classes. Thus, we introduce objectness score distillation with $\mathcal{L}_\mathrm{os-kd}$ to preserve objectness scores learned in previous steps. Moreover, in spite of the robust nature of objectness, could we further improve the objectness during continual learning? To answer this question, we propose another two distillation losses $\mathcal{L}_\mathrm{mask-kd}$ and $\mathcal{L}_\mathrm{pe-kd}$ for mask proposals and position embeddings, respectively (see \cref{fig:obj-dist} (b)). Following previous methods~\cite{liLearningForgetting2018,cermelliModelingBackgroundIncremental2020}, we distill the outputs from the last step model to mitigate the objectness forgetting on old classes.

Formally, we denote unmatched mask proposals as $\mathcal{M}^\mathrm{u} = \{m^\mathrm{u}_1, \dots, m^\mathrm{u}_{N-M}\}$ and objectness scores as $\mathcal{S}^\mathrm{u} = \{s^\mathrm{u}_1, \dots, s^\mathrm{u}_{N-M}\}$, which are the corresponding outputs from unmatched positional embeddings $\mathcal{E}_\mathrm{pos}^\mathrm{u} = \{e^\mathrm{u}_i, \dots, e^\mathrm{u}_{N-M}\}$. Additionally, the distillation process requires the corresponding outputs from the last step model, including mask proposals $\tilde{\mathcal{M}}^\mathrm{u} = \{\tilde{m}^\mathrm{u}_1, \dots, \tilde{m}^\mathrm{u}_{N-M}\}$, objectness scores $\tilde{\mathcal{S}}^\mathrm{u} = \{\tilde{s}^\mathrm{u}_1, \dots, \tilde{s}^\mathrm{u}_{N-M}\}$, and positional embeddings $\tilde{\mathcal{E}}_\mathrm{pos}^\mathrm{u} = \{\tilde{e}^\mathrm{u}_1, \dots, \tilde{e}^\mathrm{u}_{N-M}\}$. We freeze the parameters of the last step model during distillation.

\noindent\textbf{Distill objectness scores.}
We use vanilla knowledge distillation loss~\cite{hintonDistillingKnowledgeNeural2015} to mitigate the diminishing of objectness scores by minimizing Kullback-Leibler (KL) divergence between the current unmatched objectness score $\mathcal{S}^\mathrm{u}$ and the last step ones $\tilde{\mathcal{S}}^\mathrm{u}$,
\begin{equation}
    \mathcal{L}_\mathrm{os-kd} = -\frac{1}{|\mathcal{S}^\mathrm{u}|} \sum_{i=1}^{|\mathcal{S}^\mathrm{u}|} \left[\tilde{s}_{i}^\mathrm{u}\log{\frac{s_{i}^\mathrm{u}}{\tilde{s}_{i}^\mathrm{u}}}
    + (1-\tilde{s}_{i}^\mathrm{u})\log{\frac{1-s_{i}^\mathrm{u}}{1-\tilde{s}_{i}^\mathrm{u}}}\right].
\end{equation}

As objectness scores play a key role during inference,  this distillation is designed to maintain the objectness scores of previously learned mask proposals.

\noindent\textbf{Distill mask proposals.}
To preserve the knowledge of the mask proposals but also focus on proposals with high objectness scores, we reweight mask distillation with objectness scores. By distilling masks with low objectness scores, we relax the case when objectness scores fail to indicate the objectness, as we observe some mask proposals generalize to unseen classes, but their objectness scores are low. The mask distillation loss is defined as
\begin{equation}
    \mathcal{L}_\mathrm{mask-kd} = \sum_{i=1}^{|\mathcal{M}^\mathrm{u}|} \frac{\omega_i}{\sum_{i=1}^{|\mathcal{M}^\mathrm{u}|}\omega_i}\mathcal{L}_\mathrm{mask}(m_i^\mathrm{u}, \tilde{m}_i^\mathrm{u}),
\end{equation}
where $\omega_i = (\tilde{s}_{i}^\mathrm{u})^{\beta}$, and $\beta$ is a reweight strength hyperparameter.

\noindent\textbf{Distill positional embeddings.}
Similar to mask distillation, we reweight position distillation when enforcing the unmatched positional embeddings $\mathcal{E}_\mathrm{pos}^\mathrm{u}$ to be similar to the last step ones $\tilde{\mathcal{E}}_\mathrm{pos}^\mathrm{u}$. Here, we encourage high cosine similarity between the current step ones and the last step ones, which assists in preserving positional information for later multi-label class distillation,
\begin{equation}
    \mathcal{L}_\mathrm{pe-kd} = \sum_{i=1}^{|\mathcal{E}_\mathrm{pos}^\mathrm{u}|} \frac{\omega_i}{\sum_{i=1}^{|\mathcal{E}_\mathrm{pos}^\mathrm{u}|}\omega_i} \left(1 - \cos(e_i^{\mathrm{u}}, \tilde{e}_i^\mathrm{u})\right),
\end{equation}
where $\cos(\cdot)$ means cosine similarity.

Finally, we distill the objectness by minimizing $\mathcal{L}_\mathrm{obj-kd} = \mathcal{L}_\mathrm{os-kd} + \mathcal{L}_\mathrm{mask-kd} + \mathcal{L}_\mathrm{pe-kd}$.

\subsubsection{Class Distillation}
\label{sec:class-distill}

As the segmentation is decoupled to a recognition task at the second stage, the continual learning of the second stage is essentially continual multi-label classification.
We discuss class distillation in two scenarios: (i) distilling the class prediction from the last step high-objectness embeddings unmatched with GT, and (ii) distilling the class knowledge from matched embeddings (see in \cref{fig:obj-dist} (c)). During class distillation, we freeze the parameters of the last step model.

\noindent\textbf{Distill from unmatched queries.}
This scenario is similar to the pseudo-labeling of the old classes but in a soft way, minimizing the KL divergence between the class probability of the current step and the last step. Given the last step unmatched embeddings $\tilde{\mathcal{E}}_\mathrm{pos}^\mathrm{u}$, we select $n$ embeddings with high objectness score $\tilde{\mathcal{E}}_\mathrm{pos}^\mathrm{h} = \{\tilde{e}_i^\mathrm{u} \mid \tilde{s}_{i}^\mathrm{u} > \alpha \} \in \mathbb{R}^{n \times d}$, where $\alpha$ is high objectness threshold. The corresponding current step embeddings $\mathcal{E}_\mathrm{pos}^\mathrm{h}$ are also selected. After class decoding, we obtain the last step class logits $\{\tilde{z}_i^\mathrm{h} \}_{i=1}^n$ and the current ones $\{z_i^\mathrm{h} \}_{i=1}^{n}$. The knowledge distillation loss is enforced on the probability of old classes $\mathcal{C}^{1:t-1}$, \ie, $\mathcal{L}_\mathrm{cls-kd}^\mathrm{u} = \frac{1}{n} \sum_{i=1}^{n} D_\mathrm{KL}(p_{i}^\mathrm{h}\|\tilde{p}_{i}^\mathrm{h})$,
where $p_i^\mathrm{h} = \mathrm{softmax}(z^\mathrm{h}_i)$ and $\tilde{p}^\mathrm{h}_i = \mathrm{softmax}(\tilde{z}^\mathrm{h}_i)$ are the current step old class probability and the last step one decoded from $i$-th positional embedding. $D_\mathrm{KL}(\cdot)$ denotes KL divergence.

\noindent\textbf{Distill from matched queries.}
For new samples, we consider its last step model prediction as the prior knowledge to mitigate forgetting.
To extract the knowledge, we reuse the currently matched embeddings $\mathcal{E}_\mathrm{pos}^\mathrm{m}$ and get the old class probability activated by softmax $\{p_i^\mathrm{m}\}_{i=1}^{M}$. Additionally, we decode $\mathcal{E}_\mathrm{pos}^\mathrm{m}$ using the last step model,
obtaining the last step probability $\{\tilde{p}_i^\mathrm{m}\}_{i=1}^{M}$ as the target distribution. During training, we minimize matched class distillation loss $\mathcal{L}_\mathrm{cls-kd}^\mathrm{m} = \frac{1}{M} \sum_{i=1}^{M} D_\mathrm{KL}(p_{i}^\mathrm{m}\|\tilde{p}_{i}^\mathrm{m})$.

In total, we preserve class knowledge by minimizing $\mathcal{L}_\mathrm{cls-kd} = \mathcal{L}_\mathrm{cls-kd}^\mathrm{u} + \mathcal{L}_\mathrm{cls-kd}^\mathrm{m}$. Besides, we follow prior classification works~\cite{yanDynamicallyExpandableRepresentation2021,douillardDyToxTransformersContinual2022} and use an auxiliary loss $\mathcal{L}_\mathrm{cls-aux}$ with a similar purpose to previous continual segmentation methods~\cite{phanClassSimilarityWeighted2022,shangIncrementerTransformerClassIncremental2023}.
It encourages more discriminative classification of similar classes by learning to predict a merged old class and new classes (1 + $|\mathcal{C}^t|$ classes in total) at each step.
Together with segmentation loss, the loss of CoMasTRe is $\mathcal{L} = \mathcal{L}_\mathrm{seg} + \mathcal{L}_\mathrm{kd} + \mathcal{L}_\mathrm{cls-aux}$, where $\mathcal{L}_\mathrm{kd} = \mathcal{L}_\mathrm{obj-kd}+ \mathcal{L}_\mathrm{cls-kd}$.

\section{Experiments}
\label{sec:exp}

\subsection{Setup}

\noindent\textbf{Datasets.}
Following standard continual segmentation baselines~\cite{cermelliModelingBackgroundIncremental2020,douillardPLOPLearningForgetting2021,phanClassSimilarityWeighted2022}, we evaluate semantic segmentation performance of CoMasTRe on PASCAL VOC 2012~\cite{everinghamPascalVisualObject2010}~and ADE20K~\cite{zhouSceneParsingADE20K2017}. PASCAL VOC 2012 is a relatively small dataset with 20 object classes and background class, 
containing 10,582 samples for training and 1,449 for validation.
ADE20K is a large-scale semantic segmentation dataset with 150 annotated classes, containing 20,210 and 2,000 samples for training and validation.

\begin{table*}
    \centering
    \caption[voc-bench]{Comparison with previous best methods on PASCAL VOC 2012 in mIoU (\%). The highest and the second highest results are marked in \textbf{bold} and \underline{underline}. $\ast$ means results from our re-implementation.}
    \vspace{-2.0mm}
    \resizebox{\textwidth}{!}{
        \begin{tabular}{ll|cccc|cccc|cccc}
            \toprule
            \multirow{2}{*}{\textbf{Paradigm}} & \multirow{2}{*}{\textbf{Method}}                               & \multicolumn{4}{c}{\textbf{19-1} (2 tasks)} & \multicolumn{4}{c}{\textbf{15-5} (2 tasks)} & \multicolumn{4}{c}{\textbf{15-1} (6 tasks)}                                                                                                                                                                                     \\
                                               &                                                                & \textit{1-19}                               & \textit{20}                                 & \textit{all}                                & \textit{avg}      & \textit{1-15}     & \textit{16-20}    & \textit{all}      & \textit{avg}      & \textit{1-15}     & \textit{16-20}    & \textit{all}      & \textit{avg}      \\
            \midrule
            \multirow{6}{*}{Per-Pixel}         & MiB~\cite{cermelliModelingBackgroundIncremental2020}           & 71.43                                       & 23.59                                       & 69.15                                       & 73.28             & 76.37             & 49.97             & 70.08             & 75.12             & 34.22             & 13.50             & 29.29             & 54.19             \\
                                               & SDR~\cite{michieliContinualSemanticSegmentation2021}           & 68.52                                       & 23.29                                       & 66.37                                       & 71.48             & 75.21             & 46.72             & 68.64             & 74.32             & 43.08             & 19.31             & 37.42             & 54.52             \\
                                               & PLOP~\cite{douillardPLOPLearningForgetting2021}                & \underline{75.35}                           & \underline{37.35}                           & 73.54                                       & 75.47             & 75.73             & 51.71             & 70.09             & 75.19             & 65.12             & 21.11             & 54.64             & 67.21             \\
                                               & REMINDER~\cite{phanClassSimilarityWeighted2022}                & \textbf{76.48}                              & 32.34                                       & \underline{74.38}                           & \underline{76.22} & 76.11             & 50.74             & 70.07             & \underline{75.36} & 68.30             & \underline{27.23} & 58.52             & \underline{68.27} \\
                                               & RCIL~\cite{zhangRepresentationCompensationNetworks2022}        & ---                                         & ---                                         & ---                                         & ---               & \underline{78.80} & \textbf{52.00}    & \underline{72.40} & ---               & \textbf{70.60}    & 23.70             & \underline{59.40} & ---               \\
                                               & \textit{Joint}                                                 & 77.45                                       & 77.94                                       & 77.39                                       & ---               & 78.88             & 72.63             & 77.39             & ---               & 78.88             & 72.63             & 77.39             & ---               \\ \midrule
            \multirow{4}{*}{Query}             & CoMFormer$^\ast$~\cite{cermelliCoMFormerContinualLearning2023} & \underline{75.35}                           & 24.06                                       & 72.91                                       & 75.46             & 74.68             & 48.47             & 68.44             & 72.97             & 48.97             & 23.28             & 48.18             & 64.16             \\
                                               & \textit{Joint}                                                 & 77.09                                       & 71.39                                       & 76.82                                       & ---               & 78.60             & 71.12             & 76.82             & ---               & 78.60             & 71.12             & 76.82             & ---               \\ \cmidrule(l){2-14}
                                               & \textbf{CoMasTRe} (ours)                                       & 75.13                                       & \textbf{69.51}                              & \textbf{74.86}                              & \textbf{76.66}    & \textbf{79.73}    & \underline{51.93} & \textbf{73.11}    & \textbf{75.97}    & \underline{69.77} & \textbf{43.62}    & \textbf{63.54}    & \textbf{70.63}    \\
                                               & \textit{Joint}                                                 & 78.57                                       & 68.32                                       & 78.08                                       & ---               & 80.83             & 69.28             & 78.08             & ---               & 80.83             & 69.28             & 78.08             & ---               \\ \bottomrule
        \label{tab:voc}
        \end{tabular}        }
    \squeezeup
    \squeezeup
\end{table*}

\begin{table*}
    \vspace{-1.2mm}
    \centering
    \caption[ade20k-bench]{Comparison with previous methods on ADE20K in mIoU (\%). The 1\textsuperscript{st} and 2\textsuperscript{nd} highest results are marked in \textbf{bold} and \underline{underline}.}
    \vspace{-2mm}
    \resizebox{\textwidth}{!}{
        \begin{tabular}{ll|cccc|cccc|cccc}
            \toprule
            \multirow{2}{*}{\textbf{Paradigm}} & \multirow{2}{*}{\textbf{Method}}                         & \multicolumn{4}{c}{\textbf{100-50} (2 tasks)} & \multicolumn{4}{c}{\textbf{100-10} (6 tasks)} & \multicolumn{4}{c}{\textbf{100-5} (11 tasks)}                                                                                                                                                                                     \\
                                               &                                                          & \textit{1-100}                                & \textit{101-150}                              & \textit{all}                                  & \textit{avg}      & \textit{1-100}    & \textit{101-150}  & \textit{all}      & \textit{avg}      & \textit{1-100}    & \textit{101-150}  & \textit{all}      & \textit{avg}      \\ \midrule
            \multirow{6}{*}{Per-Pixel}         & MiB~\cite{cermelliModelingBackgroundIncremental2020}     & 40.50                                         & 17.20                                         & 32.80                                         & 37.30             & 38.30             & 11.30             & 29.20             & 35.10             & 36.00             & 5.70              & 26.00             & 32.70             \\
                                               & SDR~\cite{michieliContinualSemanticSegmentation2021}     & 40.52                                         & 17.17                                         & 32.79                                         & 37.31             & 37.26             & 12.13             & 28.94             & 34.48             & 33.02             & 10.63             & 25.61             & 33.07             \\
                                               & PLOP~\cite{douillardPLOPLearningForgetting2021}          & 41.87                                         & 14.89                                         & 32.94                                         & 37.39             & 40.48             & 13.61             & 31.59             & 36.64             & 35.72             & 12.18             & 27.93             & 35.10             \\
                                               & REMINDER~\cite{phanClassSimilarityWeighted2022}          & 41.55                                         & 19.16                                         & 34.14                                         & 38.43             & 38.96             & \textbf{21.28}    & \underline{33.11} & \underline{37.47} & 36.06             & \textbf{16.38}    & 29.54             & 36.49             \\
                                               & RCIL~\cite{zhangRepresentationCompensationNetworks2022}  & 42.30                                         & 18.80                                         & 34.50                                         & ---               & 39.30             & 17.60             & 32.10             & ---               & 38.50             & 11.50             & 29.60             & ---               \\
                                               & \textit{Joint}                                           & 44.34                                         & 28.21                                         & 39.00                                         & ---               & 44.34             & 28.21             & 39.00             & ---               & 44.34             & 28.21             & 39.00             & ---               \\ \midrule
            \multirow{4}{*}{Query}             & CoMFormer~\cite{cermelliCoMFormerContinualLearning2023}  & \underline{44.70}                             & \textbf{26.20}                                & \underline{38.40}                             & \underline{41.20} & \underline{40.60} & 15.60             & 32.30             & 37.40             & \underline{39.50} & 13.60             & \underline{30.90} & \underline{36.50} \\
                                               & \textit{Joint}                                           & 46.90                                         & 35.60                                         & 43.10                                         & ---               & 46.90             & 35.60             & 43.10             & ---               & 46.90             & 35.60             & 43.10             & ---               \\ \cmidrule(l){2-14}
                                               & \textbf{CoMasTRe} (ours)                                 & \textbf{45.73}                                & \underline{26.02}                             & \textbf{39.20}                                & \textbf{41.62}    & \textbf{42.32}    & \underline{18.42} & \textbf{34.41}    & \textbf{38.41}    & \textbf{40.82}    & \underline{15.83} & \textbf{32.55}    & \textbf{38.64}    \\
                                               & \textit{Joint}                                           & 48.48                                         & 36.11                                         & 44.36                                         & ---               & 48.48             & 36.11             & 44.36             & ---               & 48.48             & 36.11             & 44.36             & ---               \\ \bottomrule

        \label{tab:ade}
        \end{tabular}
    }
    \vspace{-6.5mm}
\end{table*}

\noindent\textbf{Continual Segmentation Protocols.}
Continual segmentation protocols include \textit{sequential}, \textit{disjoint}, and \textit{overlapped}, where all of them split the segmentation into $T$ steps, and each step aims to learn a unique set of classes. The \textit{sequential}  is the easiest, as the training set contains the ground truth of each pixel. In contrast, the \textit{disjoint} annotates past classes as background and excludes any training pixel belonging to future classes. The \textit{overlapped} is the most challenging and realistic one, where the dataset only contains the ground truth of current classes, but both past and future classes are annotated as background. Following previous works~\cite{douillardPLOPLearningForgetting2021,phanClassSimilarityWeighted2022,cermelliCoMFormerContinualLearning2023}, we adopt the \textit{overlapped} in benchmarks.

We evaluate CoMasTRe in 6 settings, 3 on PASCAL VOC and 3 on ADE20K. Here, we use the $B$-$I$ notation in the paper, where $B$ and $I$ denote the number of base classes and incremented classes per step. Then, the number of total steps $T = 1 + (|\mathcal{C}| - B) / I$. For example, \textit{15-1} on VOC means we start the learning process at 15 base classes, then continually train the model for 5 steps, one new class per step. In this way, the 3 settings on PASCAL VOC are (i) \textit{19-1}, 2-step learning starting with 19 classes and following 1 class, (ii) \textit{15-5}, 2-step learning starting with 15 classes and following 5 classes, and (iii) \textit{15-1}, consisting 6 learning steps with 15 classes at first and following 5 steps with one class each step. Similarly, the settings on ADE20K are \textit{100-50}, \textit{100-10}, and \textit{100-5}. 
Note that the longer the learning process, the harder it is to tackle forgetting. 

\noindent\textbf{Metric.}
We evaluate semantic segmentation performance using mean intersection over union (mIoU). In continual learning settings, the mIoU of base classes $\mathcal{C}^1$ and incremented classes $\mathcal{C}^{2:T}$ are also measured. The base class performance measures the \textit{stability} of the model, while the incremented one reflects the \textit{plasticity}. Additionally, following~\cite{douillardPLOPLearningForgetting2021,phanClassSimilarityWeighted2022,cermelliCoMFormerContinualLearning2023}, we report the average mIoU of all learning steps as \textit{avg} to evaluate the entire continual learning process. We denote \textit{joint} as the joint training of all classes.

\noindent\textbf{Implementation details.}
Following the previous continual segmentation benchmarks~\cite{cermelliModelingBackgroundIncremental2020,douillardPLOPLearningForgetting2021,cermelliCoMFormerContinualLearning2023}, we use ImageNet-pretrained ResNet101~\cite{heDeepResidualLearning2016}~as the backbone. The pixel decoder in CoMasTRe is identical to the Mask2Former~\cite{chengMaskedAttentionMaskTransformer2022} one, the mask decoder consists of $L=9$ Transformer decoder blocks, and the class decoder consists of one block. We set the number of positional queries $N$ to 20 and 100 for PASCAL VOC and ADE20K, respectively. During training, we follow the Mask2Former~\cite{chengMaskedAttentionMaskTransformer2022}~hyperparameters. We use AdamW~\cite{loshchilovDecoupledWeightDecay2018} optimizer with an initial learning rate of $1\times10^{-4}$ and weight decay of $0.05$ and apply a polynomial learning rate schedule. We train the model with a batch size of 16 for 20K iterations on PASCAL VOC and 160K iterations on ADE20K at the first learning step. In later steps, we half the learning rate to $5\times10^{-5}$ and train for 1,000 iterations per class on VOC and 500 iterations per class on ADE20K. The images are augmented with random rescaling, flipping, and color jittering then cropped to $512\times512$ resolution. We set reweight parameters $\gamma$ and $\beta$ to $2.0$. To evaluate the IoU of background during continual learning, we use the panoptic inference from Mask2Former~\cite{chengMaskedAttentionMaskTransformer2022}. During inference, we set the high objectness threshold $\alpha$ as $0.8$. Following CoMFormer~\cite{cermelliCoMFormerContinualLearning2023}, we use single-scale inference, and no replay is involved throughout training.

\subsection{Quantitative Evaluation}
We compare CoMasTRe with state-of-the-art continual segmentation methods on PASCAL VOC 2012 and ADE20K. Qualitative results are available in supplementary materials.

\noindent\textbf{PASCAL VOC 2012.} We compare our method with previous state-of-the-art methods on the 3 settings on PASCAL VOC: \textit{19-1}, \textit{15-5}, and \textit{15-1} in \cref{tab:voc}. In general, CoMasTRe significantly outperforms query-based CoMFormer~\cite{cermelliCoMFormerContinualLearning2023}~and surpasses per-pixel baselines on incremented classes by a large margin. On \textit{19-1}, CoMasTRe shows off its strong plasticity in learning new classes, with 32.16 percent points (\textit{p.p}) improvement while maintaining the knowledge of old classes. On \textit{15-5}, our model is on par with RCIL~\cite{zhangRepresentationCompensationNetworks2022}~on new classes while exceptionally great at maintaining old class performance, even surpassing the \textit{joint} upper bound of per-pixel baselines by 0.85 \textit{p.p}. \textit{15-1} is much more difficult and involves 6 learning steps. However, CoMasTRe still beats RCIL with 4.14 \textit{p.p} uplift on \textit{all} classes.
Overall, we strikes a balance between \textit{stability} and \textit{plasticity}.


\noindent\textbf{ADE20K.} As shown in \cref{tab:ade}, we report the results on ADE20K in \textit{100-50}, \textit{100-10}, and \textit{100-5}. In general, CoMasTRe outperforms previous state-of-the-art CoMFormer~\cite{cermelliCoMFormerContinualLearning2023}~in each setting by 0.8 \textit{p.p}, 2.11 \textit{p.p}, and 1.65 \textit{p.p} on \textit{all} classes, respectively. In \textit{100-50}, our method shows competitive performance with CoMFormer on incremented classes but with 0.8 \textit{p.p} gain on \textit{all} classes thanks to its higher upper bound. For longer learning processes, CoMasTRe reaches a comparable performance with REMINDER~\cite{phanClassSimilarityWeighted2022}~on new classes but significantly boosts base class mIoU (3.36 \textit{p.p} in \textit{100-10} and 4.76 \textit{p.p} in \textit{100-5}). Compared with CoMFormer, our CoMasTRe also learns better new classes with 2.82 \textit{p.p} improvement in \textit{100-10} and 2.23 \textit{p.p} in \textit{100-5} while preserving better old class knowledge, with 1.72 \textit{p.p} and 1.32 \textit{p.p} gains, respectively.

\subsection{Ablation Studies}
\noindent\textbf{Joint training results.}
To resolve scalability concerns on mask-only matching, we demonstrate the \textit{joint} performance of CoMasTRe, \ie, single-shot training on all classes. We compare the segmentation performance in mIoU on PASCAL VOC and ADE20k using ResNet50 (R50) and ResNet101 (R101) backbones in \cref{tab:joint-bench}. The results show that CoMasTRe achieves comparable performance with Mask2Former and outperforms the previous continual model CoMFormer by $\sim$1.2 \textit{p.p} on ADE20K.

\begin{table}
    \centering
    \caption[joint-bench]{Semantic segmentation results mIoU in \textit{joint} setting. Following continual settings, we use panoptic-style inference. The 1\textsuperscript{st} and 2\textsuperscript{nd} highest results are marked in \textbf{bold} and \underline{underline}.}
    \vspace{-2mm}
    \resizebox{0.85\columnwidth}{!}{
    \begin{tabular}{ll|cc}
        \toprule
        \multirow{2}{*}{\textbf{Method}}                                            & \multirow{2}{*}{\textbf{Backbone}} & \multicolumn{2}{c}{\textbf{Datasets}}                     \\
                                                                                    &                                    & VOC                                   & ADE20k            \\ \midrule
        \multirow{2}{*}{Mask2Former~\cite{chengMaskedAttentionMaskTransformer2022}} & R50                                & \textbf{79.73}                        & \underline{43.28} \\
                                                                                    & R101                               & \textbf{80.04}                        & \underline{43.51} \\ \midrule
        \multirow{2}{*}{CoMFormer~\cite{cermelliCoMFormerContinualLearning2023}}    & R50                                & 76.01                                 & 42.58             \\
                                                                                    & R101                               & 76.82                                 & 43.10             \\ \midrule
        \multirow{2}{*}{\textbf{CoMasTRe} (ours)}                                            & R50                                & \underline{77.84}                     & \textbf{43.97}    \\
                                                                                    & R101                               & \underline{78.08}                     & \textbf{44.36}    \\ \bottomrule
    \label{tab:joint-bench}
    \end{tabular}
    }
    \vspace{-4.75mm}
\end{table}

\begin{table}
    \vspace{-2mm}
    \centering
    \caption[obj-general]{Objectness transfer ability results in PASCAL VOC \textit{15-1} setting. For transfer learning, we continually train on VOC using a stage 1 mask decoder pretrained on COCO (w/o PASCAL classes). Otherwise, the model is trained on VOC only without pretraining.}
    \vspace{-2mm}
    \resizebox{0.75\columnwidth}{!}{
        \begin{tabular}{c|ccccc}
            \toprule
            \textbf{Transfer}    & \textit{1-15}  & \textit{16-20} & \textit{all}   & \textit{avg}   \\ \midrule
            \cmark               & \textbf{70.72} & \textbf{46.18} & \textbf{64.88} & \textbf{71.34} \\
            \xmark               & 69.77          & 43.62          & 63.54          & 70.63          \\
            \bottomrule
        \end{tabular}
        \label{tab:obj-trf}
        \vspace{-5mm}
    }
\end{table}


\begin{table}
    \vspace{-3mm}
    \caption[ablate-stage1]{Objectness distillation results in PASCAL VOC \textit{15-1} setting, where $\mathcal{L}_\mathrm{mask-kd}$ for mask distillation, $\mathcal{L}_\mathrm{os-kd}$ for objectness score distillation, $\mathcal{L}_\mathrm{pe-kd}$ for position distillation.}
    \vspace{-2mm}
    \resizebox{\columnwidth}{!}{
        \begin{tabular}{ccc|cccc}
            \toprule
            $\mathcal{L}_\mathrm{mask-kd}$ & $\mathcal{L}_\mathrm{os-kd}$ & $\mathcal{L}_\mathrm{pe-kd}$ & \textit{1-15}  & \textit{16-20} & \textit{all}   & \textit{avg}   \\ \midrule
            \cmark                     &                              &                              & 13.14          & 35.42          & 18.44          & 47.43          \\
                                           & \cmark                   &                              & 63.75          & 37.97          & 57.61          & 62.72          \\
            \cmark                     & \cmark                   &                              & 66.23          & 38.57          & 59.64          & 64.61          \\
            \cmark                     & \cmark                   & \cmark                   & \textbf{69.77} & \textbf{43.62} & \textbf{63.54} & \textbf{70.63} \\ \bottomrule
        \end{tabular}
    }
    \vspace{-4mm}
    \label{tab:ablate-stage1}
\end{table}

\noindent\textbf{Objectness transfer ability analysis.}
\label{para:obj-trf}
To study the transfer ability of objectness, we first pretrain a stage 1 mask decoder on modified COCO 2017~\cite{linMicrosoftCOCOCommon2014}, where instances of the classes appearing in PASCAL VOC are removed to prevent information leaks. Then, we transfer the stage 1 and continually train it on 15 base classes of VOC with objectness distillation. Finally, we follow standard procedures by training 1 class per step. As shown in~\cref{tab:obj-trf}, compared with the model trained on VOC only, we get 1.34 \textit{p.p} performance uplift on \textit{all} metric in PASCAL VOC \textit{15-1} setting, indicating the strong transfer ability of objectness.

\noindent\textbf{Effectiveness of objectness distillation.}
We ablate objectness distillation components with optimal configuration on other parts. The ablation is conducted in four cases: (1) $\mathcal{L}_\mathrm{mask-kd}$ for mask only, (2) $\mathcal{L}_\mathrm{os-kd}$ for objectness score only, (3) $\mathcal{L}_\mathrm{mask-kd} + \mathcal{L}_\mathrm{os-kd}$ for mask and objectness score, and (4) $\mathcal{L}_\mathrm{mask-kd} + \mathcal{L}_\mathrm{os-kd} + \mathcal{L}_\mathrm{pe-kd}$ for mask, objectness score, and position.
As shown in \cref{tab:ablate-stage1}, we report the performance in PASCAL VOC \textit{15-1} setting. By comparing case 1 and 3, without $\mathcal{L}_\mathrm{os-kd}$, base class mIoU degrades significantly as the base objectness score diminishes.
By comparing case 2 and 3, we observe slight forgetting of mask proposals without $\mathcal{L}_\mathrm{mask-kd}$ ($\sim$2 \textit{p.p} drop on \textit{all} metric), showing the forgetting robustness of mask proposals.
The position distillation aims for better alignment of localized information and leads to better classification results, with $\sim$4 \textit{p.p} \textit{all} performance gain when comparing case 3 and 4.

\noindent\textbf{Effectiveness of class distillation.}
We perform ablation in VOC \textit{15-1} setting by analyzing the effectiveness of $\mathcal{L}^\mathrm{u}_\mathrm{cls-kd}$ and $\mathcal{L}^\mathrm{m}_\mathrm{cls-kd}$ in \cref{sec:class-distill}. The results in \cref{tab:ablate-stage2-distill} indicate $\mathcal{L}^\mathrm{u}_\mathrm{cls-kd}$ playing a pivotal role in alleviating forgetting, coinciding with pseudo-labeling in previous methods~\cite{douillardPLOPLearningForgetting2021,cermelliCoMFormerContinualLearning2023}. With $\mathcal{L}^\mathrm{m}_\mathrm{cls-kd}$, the forgetting issue is mitigated even further.

\begin{table}
    \centering
    \caption[ablate-stage2-distill]{Ablation results on class distillation in VOC \textit{15-1} setting.}
    \vspace{-2mm}
    \resizebox{0.88\columnwidth}{!}{
        \begin{tabular}{cc|cccc}
            \toprule
            $\mathcal{L}^\mathrm{m}_\mathrm{cls-kd}$ & $\mathcal{L}^\mathrm{u}_\mathrm{cls-kd}$ & \textit{1-15}  & \textit{16-20} & \textit{all}   & \textit{avg}   \\ \midrule
            \cmark                               &                                      & 34.63          & 42.36          & 36.47          & 53.92          \\
                                                 & \cmark                               & 65.24          & 39.53          & 59.12          & 64.23          \\
            \cmark                               & \cmark                               & \textbf{69.77} & \textbf{43.62} & \textbf{63.54} & \textbf{70.63} \\ \bottomrule
        \end{tabular}
    }
    \label{tab:ablate-stage2-distill}
    \vspace{-1.5mm}
\end{table}

\begin{table}
    \centering
    \vspace{-1mm}
    \caption[ablate-stage2-cls]{More ablation results on stage 2 components in PASCAL VOC \textit{15-1} setting, where ``TQ'' denotes task queries, ``Aux'' denotes auxiliary loss $\mathcal{L}_\mathrm{cls-aux}$, and ``Focal'' denotes focal loss $\mathcal{L}_\mathrm{cls}$.}
    \vspace{-2mm}
    \resizebox{0.9\columnwidth}{!}{
        \begin{tabular}{ccc|cccc}
            \toprule
            TQ         & Aux        & Focal      & \textit{1-15}  & \textit{16-20} & \textit{all}   & \textit{avg}   \\ \midrule
                       &            &            & 59.30          & 25.96          & 51.36          & 59.44          \\
            \cmark &            &            & 63.66          & 29.32          & 55.48          & 62.45          \\
                       & \cmark &            & 65.21          & 35.53          & 58.14          & 63.93          \\
                       &            & \cmark & 66.43 	       & 40.20          & 60.18          & 64.72          \\
            \cmark & \cmark &            & 67.91          & 38.41          & 60.89          & 65.06          \\
            \cmark & \cmark & \cmark & \textbf{69.77} & \textbf{43.62} & \textbf{63.54} & \textbf{70.63} \\
            \bottomrule
        \end{tabular}
        }
    \vspace{-3mm}
    \label{tab:ablate-stage2-arch}
\end{table}

\noindent\textbf{Effectiveness of stage 2 other components.}
As shown in~\cref{tab:ablate-stage2-arch}, under optimal distillation configurations, we investigate the contribution of stage 2 other components, including task queries (TQ), focal loss $\mathcal{L}_\mathrm{cls}$ (Focal) in \cref{para:stage-2}, and auxiliary loss $\mathcal{L}_\mathrm{cls-aux}$ (Aux) in \cref{sec:class-distill}. Note that when not using task queries, one query is used across all tasks; when not using focal loss, cross-entropy is applied. We first ablate the improvements with each improvement only. First, with task queries, we get $\sim$4 \textit{p.p} improvement on \textit{all} metric as they reduce the learning interference between tasks. Second, the auxiliary loss assists in learning without forgetting new classes similar to old classes by increasing $\sim$7 \textit{p.p} of \textit{all} metric. Third, when using the focal loss to smooth the class probability distribution, it facilitates the distillation process and results in $\sim$9 \textit{p.p} mIoU gain. With all the improvements, we boost the final performance by $\sim$12 \textit{p.p}.

\section{Conclusion}
\label{sec:conclusion}

In this paper, we present CoMasTRe, a continual segmentation framework by disentangling the challenging segmentation problem to objectness learning and class recognition. To leverage the properties of objectness, we propose a two-stage query-based segmenter and distill objectness and classification knowledge separately to alleviate forgetting. Extensive experiments show that our method achieves considerably better performance compared with state-of-the-art methods on PASCAL VOC 2012 and ADE20K. For future works, CoMasTRe provides a decoupled way to tackle continual semantic segmentation and could be extended to continual panoptic and instance segmentation.



{
    \small
    \vspace{1mm}
    \noindent\textbf{Acknowledgments.}
This work was supported by the National Key R\&D Program of China (No. 2022YFE0200300), the National Natural Science Foundation of China (No. 61972323, 62331003), Suzhou Basic Research Program (SYG202316) and XJTLU REF-22-01-010, XJTLU AI University Research Centre, Jiangsu Province Engineering Research Centre of Data Science and Cognitive Computation at XJTLU and SIP AI innovation platform (YZCXPT2022103).

   \bibliographystyle{ieeenat_fullname}
   \bibliography{main}

\begin{thebibliography}{66}
\providecommand{\natexlab}[1]{#1}
\providecommand{\url}[1]{\texttt{#1}}
\expandafter\ifx\csname urlstyle\endcsname\relax
  \providecommand{\doi}[1]{doi: #1}\else
  \providecommand{\doi}{doi: \begingroup \urlstyle{rm}\Url}\fi

\bibitem[Buzzega et~al.(2020)Buzzega, Boschini, Porrello, Abati, and Calderara]{buzzegaDarkExperienceGeneral2020}
Pietro Buzzega, Matteo Boschini, Angelo Porrello, Davide Abati, and Simone Calderara.
\newblock Dark {{Experience}} for {{General Continual Learning}}: A {{Strong}}, {{Simple Baseline}}.
\newblock In \emph{NeurIPS}, 2020.

\bibitem[Carion et~al.(2020)Carion, Massa, Synnaeve, Usunier, Kirillov, and Zagoruyko]{carionEndtoEndObjectDetection2020a}
Nicolas Carion, Francisco Massa, Gabriel Synnaeve, Nicolas Usunier, Alexander Kirillov, and Sergey Zagoruyko.
\newblock End-to-{{End Object Detection}} with {{Transformers}}.
\newblock In \emph{ECCV}, 2020.

\bibitem[Cermelli et~al.(2020)Cermelli, Mancini, Bulo, Ricci, and Caputo]{cermelliModelingBackgroundIncremental2020}
Fabio Cermelli, Massimiliano Mancini, Samuel~Rota Bulo, Elisa Ricci, and Barbara Caputo.
\newblock Modeling the {{Background}} for {{Incremental Learning}} in {{Semantic Segmentation}}.
\newblock In \emph{CVPR}, 2020.

\bibitem[Cermelli et~al.(2022)Cermelli, Fontanel, Tavera, Ciccone, and Caputo]{cermelliIncrementalLearningSemantic2022}
Fabio Cermelli, Dario Fontanel, Antonio Tavera, Marco Ciccone, and Barbara Caputo.
\newblock Incremental {{Learning}} in {{Semantic Segmentation}} from {{Image Labels}}.
\newblock In \emph{CVPR}, 2022.

\bibitem[Cermelli et~al.(2023)Cermelli, Cord, and Douillard]{cermelliCoMFormerContinualLearning2023}
Fabio Cermelli, Matthieu Cord, and Arthur Douillard.
\newblock {{CoMFormer}}: {{Continual Learning}} in {{Semantic}} and {{Panoptic Segmentation}}.
\newblock In \emph{CVPR}, 2023.

\bibitem[Cha et~al.(2021{\natexlab{a}})Cha, Lee, and Shin]{chaCo2LContrastiveContinual2021}
Hyuntak Cha, Jaeho Lee, and Jinwoo Shin.
\newblock {{Co2L}}: {{Contrastive Continual Learning}}.
\newblock In \emph{ICCV}, 2021{\natexlab{a}}.

\bibitem[Cha et~al.(2021{\natexlab{b}})Cha, {Kim}, Yoo, and Moon]{chaSSULSemanticSegmentation2021}
Sungmin Cha, Beomyoung {Kim}, YoungJoon Yoo, and Taesup Moon.
\newblock {{SSUL}}: {{Semantic Segmentation}} with {{Unknown Label}} for {{Exemplar-based Class-Incremental Learning}}.
\newblock In \emph{NeurIPS}, 2021{\natexlab{b}}.

\bibitem[Chen et~al.(2017)Chen, Papandreou, Schroff, and Adam]{chenRethinkingAtrousConvolution2017}
Liang-Chieh Chen, George Papandreou, Florian Schroff, and Hartwig Adam.
\newblock Rethinking {{Atrous Convolution}} for {{Semantic Image Segmentation}}.
\newblock \emph{arXiv preprint arXiv:1706.05587}, 2017.

\bibitem[Cheng et~al.(2021)Cheng, Schwing, and Kirillov]{chengPerPixelClassificationNot2021}
Bowen Cheng, Alex Schwing, and Alexander Kirillov.
\newblock Per-{{Pixel Classification}} is {{Not All You Need}} for {{Semantic Segmentation}}.
\newblock In \emph{NeurIPS}, 2021.

\bibitem[Cheng et~al.(2022)Cheng, Misra, Schwing, Kirillov, and Girdhar]{chengMaskedAttentionMaskTransformer2022}
Bowen Cheng, Ishan Misra, Alexander~G. Schwing, Alexander Kirillov, and Rohit Girdhar.
\newblock Masked-{{Attention Mask Transformer}} for {{Universal Image Segmentation}}.
\newblock In \emph{CVPR}, 2022.

\bibitem[Douillard et~al.(2020)Douillard, Cord, Ollion, Robert, and Valle]{douillardPODNetPooledOutputs2020}
Arthur Douillard, Matthieu Cord, Charles Ollion, Thomas Robert, and Eduardo Valle.
\newblock {{PODNet}}: {{Pooled Outputs Distillation}} for {{Small-Tasks Incremental Learning}}.
\newblock In \emph{ECCV}, 2020.

\bibitem[Douillard et~al.(2021)Douillard, Chen, Dapogny, and Cord]{douillardPLOPLearningForgetting2021}
Arthur Douillard, Yifu Chen, Arnaud Dapogny, and Matthieu Cord.
\newblock {{PLOP}}: {{Learning}} without {{Forgetting}} for {{Continual Semantic Segmentation}}.
\newblock In \emph{CVPR}, 2021.

\bibitem[Douillard et~al.(2022)Douillard, Ram{\'e}, Couairon, and Cord]{douillardDyToxTransformersContinual2022}
Arthur Douillard, Alexandre Ram{\'e}, Guillaume Couairon, and Matthieu Cord.
\newblock {{DyTox}}: {{Transformers}} for {{Continual Learning with DYnamic TOken eXpansion}}.
\newblock In \emph{CVPR}, 2022.

\bibitem[Everingham et~al.(2010)Everingham, Van~Gool, Williams, Winn, and Zisserman]{everinghamPascalVisualObject2010}
Mark Everingham, Luc Van~Gool, Christopher K.~I. Williams, John Winn, and Andrew Zisserman.
\newblock The {{Pascal Visual Object Classes}} ({{VOC}}) {{Challenge}}.
\newblock \emph{IJCV}, 88\penalty0 (2):\penalty0 303--338, 2010.

\bibitem[Fagot and Cook(2006)]{Fagot2006EvidenceFL}
Jo{\"e}l Fagot and Robert~G. Cook.
\newblock Evidence for {{Large Long-Term Memory Capacities}} in {{Baboons}} and {{Pigeons}} and its {{Implications}} for {{Learning}} and the {{Evolution}} of {{Cognition}}.
\newblock \emph{PNAS}, 103\penalty0 (46):\penalty0 17564--17567, 2006.

\bibitem[French(1999)]{frenchCatastrophicForgettingConnectionist1999}
Robert~M. French.
\newblock {{Catastrophic Forgetting}} in {{Connectionist Networks}}.
\newblock \emph{TiCS}, 3\penalty0 (4):\penalty0 128--135, 1999.

\bibitem[Grutzendler et~al.(2002)Grutzendler, Kasthuri, and Gan]{Grutzendler2002LongtermDS}
Jaime Grutzendler, Narayanan Kasthuri, and Wen-Biao Gan.
\newblock {{Long-Term Dendritic Spine Stability}} in the {{Adult Cortex}}.
\newblock \emph{Nature}, 420:\penalty0 812--816, 2002.

\bibitem[He et~al.(2016)He, Zhang, Ren, and Sun]{heDeepResidualLearning2016}
Kaiming He, Xiangyu Zhang, Shaoqing Ren, and Jian Sun.
\newblock Deep {{Residual Learning}} for {{Image Recognition}}.
\newblock In \emph{CVPR}, 2016.

\bibitem[Hinton et~al.(2015)Hinton, Vinyals, and Dean]{hintonDistillingKnowledgeNeural2015}
Geoffrey Hinton, Oriol Vinyals, and Jeff Dean.
\newblock Distilling the {{Knowledge}} in a {{Neural Network}}.
\newblock \emph{arXiv preprint arXiv:1503.02531}, 2015.

\bibitem[Hsieh et~al.(2023)Hsieh, Chen, Cai, Wei, Yang, and Chen]{hsiehClassincrementalContinualLearning2023}
Yu-Hsing Hsieh, Guan-Sheng Chen, Shun-Xian Cai, Ting-Yun Wei, Huei-Fang Yang, and Chu-Song Chen.
\newblock Class-incremental {{Continual Learning}} for {{Instance Segmentation}} with {{Image-level Weak Supervision}}.
\newblock In \emph{ICCV}, 2023.

\bibitem[Jain et~al.(2023)Jain, Li, Chiu, Hassani, Orlov, and Shi]{jainOneFormerOneTransformer2023}
Jitesh Jain, Jiachen Li, Mang~Tik Chiu, Ali Hassani, Nikita Orlov, and Humphrey Shi.
\newblock {{OneFormer}}: {{One Transformer}} to {{Rule Universal Image Segmentation}}.
\newblock In \emph{CVPR}, 2023.

\bibitem[Kirkpatrick et~al.(2017)Kirkpatrick, Pascanu, Rabinowitz, Veness, Desjardins, Rusu, Milan, Quan, Ramalho, {Grabska-Barwinska}, Hassabis, Clopath, Kumaran, and Hadsell]{kirkpatrickOvercomingCatastrophicForgetting2017}
James Kirkpatrick, Razvan Pascanu, Neil Rabinowitz, Joel Veness, Guillaume Desjardins, Andrei Rusu, Kieran Milan, John Quan, Tiago Ramalho, Agnieszka {Grabska-Barwinska}, Demis Hassabis, Claudia Clopath, Dharshan Kumaran, and Raia Hadsell.
\newblock {{Overcoming Catastrophic Forgetting}} in {{Neural Networks}}.
\newblock \emph{PNAS}, 114\penalty0 (13):\penalty0 3521--3526, 2017.

\bibitem[Kuhn(1955)]{kuhnHungarianMethod}
Harold~W. Kuhn.
\newblock The {{Hungarian Method}} for the {{Assignment Problem}}.
\newblock \emph{Nav. Res. Logist.}, 2\penalty0 (1-2):\penalty0 83--97, 1955.

\bibitem[Li et~al.(2022)Li, Zhang, Liu, Guo, Ni, and Zhang]{liDNDETRAccelerateDETR2022}
Feng Li, Hao Zhang, Shilong Liu, Jian Guo, Lionel~M. Ni, and Lei Zhang.
\newblock {{DN-DETR}}: {{Accelerate DETR Training}} by {{Introducing Query DeNoising}}.
\newblock In \emph{CVPR}, 2022.

\bibitem[Li and Hoiem(2018)]{liLearningForgetting2018}
Zhizhong Li and Derek Hoiem.
\newblock Learning without {{Forgetting}}.
\newblock \emph{IEEE TPAMI}, 40\penalty0 (12):\penalty0 2935--2947, 2018.

\bibitem[Lin et~al.(2014)Lin, Maire, Belongie, Hays, Perona, Ramanan, Doll{\'a}r, and Zitnick]{linMicrosoftCOCOCommon2014}
Tsung-Yi Lin, Michael Maire, Serge Belongie, James Hays, Pietro Perona, Deva Ramanan, Piotr Doll{\'a}r, and C.~Lawrence Zitnick.
\newblock Microsoft {{COCO}}: {{Common Objects}} in {{Context}}.
\newblock In \emph{ECCV}, 2014.

\bibitem[Lin et~al.(2017)Lin, Goyal, Girshick, He, and Dollar]{linFocalLossDense2017}
Tsung-Yi Lin, Priya Goyal, Ross Girshick, Kaiming He, and Piotr Dollar.
\newblock Focal {{Loss}} for {{Dense Object Detection}}.
\newblock In \emph{CVPR}, 2017.

\bibitem[Liu et~al.(2022)Liu, Zhang, Bai, Zhang, and Zhao]{9656684}
Man Liu, Chunjie Zhang, Huihui Bai, Riquan Zhang, and Yao Zhao.
\newblock {{Cross-Part Learning}} for {{Fine-Grained Image Classification}}.
\newblock \emph{IEEE TIP}, 31:\penalty0 748--758, 2022.

\bibitem[Liu et~al.(2023)Liu, Li, Zhang, Wei, Bai, and Zhao]{Liu2023ProgressiveSM}
Man Liu, Feng Li, Chunjie Zhang, Yunchao Wei, Huihui Bai, and Yao Zhao.
\newblock {{Progressive Semantic-Visual Mutual Adapti-on}} for {{Generalized Zero-shot Learning}}.
\newblock In \emph{CVPR}, 2023.

\bibitem[Loshchilov and Hutter(2018)]{loshchilovDecoupledWeightDecay2018}
Ilya Loshchilov and Frank Hutter.
\newblock Decoupled {{Weight Decay Regularization}}.
\newblock In \emph{ICLR}, 2018.

\bibitem[McCloskey and Cohen(1989)]{mccloskeyCatastrophicInterferenceConnectionist1989}
Michael McCloskey and Neal~J. Cohen.
\newblock Catastrophic {{Interference}} in {{Connectionist Networks}}: {{The Sequential Learning Problem}}.
\newblock In \emph{Psychol. Learn. Motiv.}, pages 109--165. 1989.

\bibitem[Michieli and Zanuttigh(2021)]{michieliContinualSemanticSegmentation2021}
Umberto Michieli and Pietro Zanuttigh.
\newblock Continual {{Semantic Segmentation}} via {{Repulsion-Attraction}} of {{Sparse}} and {{Disentangled Latent Representations}}.
\newblock In \emph{CVPR}, 2021.

\bibitem[Mukhoti et~al.(2020)Mukhoti, Kulharia, Sanyal, Golodetz, Torr, and Dokania]{mukhotiCalibratingDeepNeural2020}
Jishnu Mukhoti, Viveka Kulharia, Amartya Sanyal, Stuart Golodetz, Philip Torr, and Puneet Dokania.
\newblock Calibrating {{Deep Neural Networks}} using {{Focal Loss}}.
\newblock In \emph{NeurIPS}, 2020.

\bibitem[Neuhold et~al.(2017)Neuhold, Ollmann, Rota~Bulo, and Kontschieder]{neuholdMapillaryVistasDataset2017}
Gerhard Neuhold, Tobias Ollmann, Samuel Rota~Bulo, and Peter Kontschieder.
\newblock The {{Mapillary Vistas Dataset}} for {{Semantic Understanding}} of {{Street Scenes}}.
\newblock In \emph{ICCV}, 2017.

\bibitem[Ozdemir and Goksel(2019)]{ozdemirExtendingPretrainedSegmentation2019}
Firat Ozdemir and Orcun Goksel.
\newblock {{Extending Pretrained Segmentation Networks}} with {{Additional Anatomical Structures}}.
\newblock \emph{IJCARS}, 14\penalty0 (7):\penalty0 1187--1195, 2019.

\bibitem[Phan et~al.(2022)Phan, Ta, Phung, {Tran-Thanh}, and Bouzerdoum]{phanClassSimilarityWeighted2022}
Minh~Hieu Phan, The-Anh Ta, Son~Lam Phung, Long {Tran-Thanh}, and Abdesselam Bouzerdoum.
\newblock Class {{Similarity Weighted Knowledge Distillation}} for {{Continual Semantic Segmentation}}.
\newblock In \emph{CVPR}, 2022.

\bibitem[Razdaibiedina et~al.(2023)Razdaibiedina, Mao, Hou, Khabsa, Lewis, and Almahairi]{razdaibiedinaProgressivePromptsContinual2023}
Anastasia Razdaibiedina, Yuning Mao, Rui Hou, Madian Khabsa, Mike Lewis, and Amjad Almahairi.
\newblock Progressive {{Prompts}}: {{Continual Learning}} for {{Language Models}} without {{Forgetting}}.
\newblock In \emph{ICLR}, 2023.

\bibitem[Riemer et~al.(2019)Riemer, Cases, Ajemian, Liu, Rish, Tu, and Tesauro]{riemerLearningLearnForgetting2019}
Matthew Riemer, Ignacio Cases, Robert Ajemian, Miao Liu, Irina Rish, Yuhai Tu, and {and}~Gerald Tesauro.
\newblock Learning to {{Learn}} without {{Forgetting}} by {{Maximizing Transfer}} and {{Minimizing Interference}}.
\newblock In \emph{ICLR}, 2019.

\bibitem[Robins(1995)]{robinsCatastrophicForgettingRehearsal1995}
Anthony Robins.
\newblock Catastrophic {{Forgetting}}, {{Rehearsal}} and {{Pseudorehearsal}}.
\newblock \emph{Connect. Sci.}, 7\penalty0 (2):\penalty0 123--146, 1995.

\bibitem[Shang et~al.(2023)Shang, Li, Meng, Wu, Qiu, and Wang]{shangIncrementerTransformerClassIncremental2023}
Chao Shang, Hongliang Li, Fanman Meng, Qingbo Wu, He-qian Qiu, and Lanxiao Wang.
\newblock Incrementer: {{Transformer}} for {{Class-Incremental Semantic Segmentation}} with {{Knowledge Distillation Focusing}} on {{Old Class}}.
\newblock In \emph{CVPR}, 2023.

\bibitem[Smith et~al.(2023)Smith, Karlinsky, Gutta, Cascante-Bonilla, Kim, Arbelle, Panda, Feris, and Kira]{smithCODAPromptCOntinualDecomposed2023}
James~Seale Smith, Leonid Karlinsky, Vyshnavi Gutta, Paola Cascante-Bonilla, Donghyun Kim, Assaf Arbelle, Rameswar Panda, Rogerio Feris, and Zsolt Kira.
\newblock {{CODA-Prompt}}: {{COntinual Decomposed Attention-based Prompting}} for {{Rehearsal-Free Continual Learning}}.
\newblock In \emph{CVPR}, 2023.

\bibitem[Strudel et~al.(2021)Strudel, Garcia, Laptev, and Schmid]{strudelSegmenterTransformerSemantic2021}
Robin Strudel, Ricardo Garcia, Ivan Laptev, and Cordelia Schmid.
\newblock Segmenter: {{Transformer}} for {{Semantic Segmentation}}.
\newblock In \emph{ICCV}, 2021.

\bibitem[Sudre et~al.(2017)Sudre, Li, Vercauteren, Ourselin, and Jorge~Cardoso]{sudreGeneralisedDiceOverlap2017}
Carole~H. Sudre, Wenqi Li, Tom Vercauteren, Sebastien Ourselin, and M. Jorge~Cardoso.
\newblock Generalised {{Dice Overlap}} as a {{Deep Learning Loss Function}} for {{Highly Unbalanced Segmentations}}.
\newblock In \emph{DLMIA}, 2017.

\bibitem[Vaswani et~al.(2017)Vaswani, Shazeer, Parmar, Uszkoreit, Jones, Gomez, Kaiser, and Polosukhin]{vaswaniAttentionAllYou2017}
Ashish Vaswani, Noam Shazeer, Niki Parmar, Jakob Uszkoreit, Llion Jones, Aidan~N Gomez, {\L}ukasz Kaiser, and Illia Polosukhin.
\newblock Attention is {{All}} you {{Need}}.
\newblock In \emph{NeurIPS}, 2017.

\bibitem[Wang et~al.(2022{\natexlab{a}})Wang, Zhou, Ye, and Zhan]{wangFOSTERFeatureBoosting2022}
Fu-Yun Wang, Da-Wei Zhou, Han-Jia Ye, and De-Chuan Zhan.
\newblock {{FOSTER}}: {{Feature Boosting}} and {{Compression}} for {{Class-Incremental Learning}}.
\newblock In \emph{ECCV}, 2022{\natexlab{a}}.

\bibitem[Wang et~al.(2020)Wang, Zhu, Green, Adam, Yuille, and Chen]{wangAxialDeepLabStandAloneAxialAttention2020}
Huiyu Wang, Yukun Zhu, Bradley Green, Hartwig Adam, Alan Yuille, and Liang-Chieh Chen.
\newblock Axial-{{DeepLab}}: {{Stand-Alone Axial-Attention}} for {{Panoptic Segmentation}}.
\newblock In \emph{ECCV}, 2020.

\bibitem[Wang et~al.(2021)Wang, Zhu, Adam, Yuille, and Chen]{wangMaXDeepLabEndtoEndPanoptic2021}
Huiyu Wang, Yukun Zhu, Hartwig Adam, Alan Yuille, and Liang-Chieh Chen.
\newblock {{MaX-DeepLab}}: {{End-to-End Panoptic Segmentation}} with {{Mask Transformers}}.
\newblock In \emph{CVPR}, 2021.

\bibitem[Wang et~al.(2023)Wang, Zhang, Yu, and Xiao]{Wang2023HuntingSD}
Xiaoyang Wang, Bingfeng Zhang, Limin Yu, and Jimin Xiao.
\newblock {{Hunting Sparsity}}: {{Density-Guided Contrastive Learning}} for {{Semi-Supervised Semantic Segmentation}}.
\newblock In \emph{CVPR}, 2023.

\bibitem[Wang et~al.(2022{\natexlab{b}})Wang, Zhang, Ebrahimi, Sun, Zhang, Lee, Ren, Su, Perot, Dy, and Pfister]{wangDualPromptComplementaryPrompting2022}
Zifeng Wang, Zizhao Zhang, Sayna Ebrahimi, Ruoxi Sun, Han Zhang, Chen-Yu Lee, Xiaoqi Ren, Guolong Su, Vincent Perot, Jennifer Dy, and Tomas Pfister.
\newblock {{DualPrompt}}: {{Complementary Prompting}} for~{{Rehearsal-Free Continual Learning}}.
\newblock In \emph{ECCV}, 2022{\natexlab{b}}.

\bibitem[Wang et~al.(2022{\natexlab{c}})Wang, Zhang, Lee, Zhang, Sun, Ren, Su, Perot, Dy, and Pfister]{wangLearningPromptContinual2022}
Zifeng Wang, Zizhao Zhang, Chen-Yu Lee, Han Zhang, Ruoxi Sun, Xiaoqi Ren, Guolong Su, Vincent Perot, Jennifer Dy, and Tomas Pfister.
\newblock Learning {{To Prompt}} for {{Continual Learning}}.
\newblock In \emph{CVPR}, 2022{\natexlab{c}}.

\bibitem[Xiao et~al.(2023)Xiao, Zhang, Feng, Liu, {van de Weijer}, and Cheng]{xiaoEndpointsWeightFusion2023}
Jiawen Xiao, Changbin Zhang, Jiekang Feng, Xialei Liu, Joost {van de Weijer}, and Mingming Cheng.
\newblock Endpoints {{Weight Fusion}} for {{Class Incremental Semantic Segmentation}}.
\newblock In \emph{CVPR}, 2023.

\bibitem[Xiao et~al.(2018)Xiao, Liu, Zhou, Jiang, and Sun]{xiaoUnifiedPerceptualParsing2018}
Tete Xiao, Yingcheng Liu, Bolei Zhou, Yuning Jiang, and Jian Sun.
\newblock Unified {{Perceptual Parsing}} for {{Scene Understanding}}.
\newblock In \emph{ECCV}, 2018.

\bibitem[Xie et~al.(2021)Xie, Wang, Yu, Anandkumar, Alvarez, and Luo]{xieSegFormerSimpleEfficient2021}
Enze Xie, Wenhai Wang, Zhiding Yu, Anima Anandkumar, Jose~M. Alvarez, and Ping Luo.
\newblock {{SegFormer}}: {{Simple}} and {{Efficient Design}} for {{Semantic Segmentation}} with {{Transformers}}.
\newblock In \emph{NeurIPS}, 2021.

\bibitem[Yan et~al.(2021)Yan, Xie, and He]{yanDynamicallyExpandableRepresentation2021}
Shipeng Yan, Jiangwei Xie, and Xuming He.
\newblock {{DER}}: {{Dynamically Expandable Representation}} for {{Class Incremental Learning}}.
\newblock In \emph{CVPR}, 2021.

\bibitem[Yu et~al.(2022)Yu, Wang, Qiao, Collins, Zhu, Adam, Yuille, and Chen]{yuKmeansMaskTransformer2022}
Qihang Yu, Huiyu Wang, Siyuan Qiao, Maxwell Collins, Yukun Zhu, Hartwig Adam, Alan Yuille, and Liang-Chieh Chen.
\newblock K-means {{Mask Transformer}}.
\newblock In \emph{ECCV}, 2022.

\bibitem[Zhang et~al.(2023{\natexlab{a}})Zhang, Xiao, Wei, and Zhao]{Zhang2023CredibleDL}
Bingfeng Zhang, Jimin Xiao, Yunchao Wei, and Yao Zhao.
\newblock {{Credible Dual-Expert Learning}} for {{Weakly Supervised Semantic Segmentation}}.
\newblock \emph{IJCV}, 131:\penalty0 1892--1908, 2023{\natexlab{a}}.

\bibitem[Zhang et~al.(2022{\natexlab{a}})Zhang, Xiao, Liu, Chen, and Cheng]{zhangRepresentationCompensationNetworks2022}
Changbin Zhang, Jiawen Xiao, Xialei Liu, Yingcong Chen, and Mingming Cheng.
\newblock Representation {{Compensation Networks}} for {{Continual Semantic Segmentation}}.
\newblock In \emph{CVPR}, 2022{\natexlab{a}}.

\bibitem[Zhang et~al.(2023{\natexlab{b}})Zhang, Wang, Kang, Chen, and Wei]{Zhang_2023_ICCV}
Gengwei Zhang, Liyuan Wang, Guoliang Kang, Ling Chen, and Yunchao Wei.
\newblock {{SLCA}}: {{Slow Learner}} with {{Classifier Alignment}} for {{Continual Learning}} on a {{Pre-trained Model}}.
\newblock In \emph{ICCV}, 2023{\natexlab{b}}.

\bibitem[Zhang et~al.(2023{\natexlab{c}})Zhang, Li, Liu, Zhang, Su, Zhu, Ni, and Shum]{zhangDINODETRImproved2023}
Hao Zhang, Feng Li, Shilong Liu, Lei Zhang, Hang Su, Jun Zhu, Lionel Ni, and Heung-Yeung Shum.
\newblock {{DINO}}: {{DETR}} with {{Improved DeNoising Anchor Boxes}} for {{End-to-End Object Detection}}.
\newblock In \emph{ICLR}, 2023{\natexlab{c}}.

\bibitem[Zhang et~al.(2022{\natexlab{b}})Zhang, Gao, Fang, Jiao, and Wei]{zhangMiningUnseenClasses2022}
Zekang Zhang, Guangyu Gao, Zhiyuan Fang, Jianbo Jiao, and Yunchao Wei.
\newblock Mining {{Unseen Classes}} via {{Regional Objectness}}: {{A Simple Baseline}} for {{Incremental Segmentation}}.
\newblock In \emph{NeurIPS}, 2022{\natexlab{b}}.

\bibitem[Zhang et~al.(2023{\natexlab{d}})Zhang, Gao, Jiao, Liu, and Wei]{zhangCoinSegContrastInter2023}
Zekang Zhang, Guangyu Gao, Jianbo Jiao, Chi~Harold Liu, and Yunchao Wei.
\newblock {{CoinSeg}}: {{Contrast Inter-}} and {{Intra- Class Representations}} for {{Incremental Segmentation}}.
\newblock In \emph{ICCV}, 2023{\natexlab{d}}.

\bibitem[Zhao et~al.(2024)Zhao, Tang, Wang, and Xiao]{Zhao2024SFCSF}
Xinqiao Zhao, Feilong Tang, Xiaoyang Wang, and Jimin Xiao.
\newblock {{SFC}}: {{Shared Feature Calibration}} in {{Weakly Supervised Semantic Segmentation}}.
\newblock In \emph{AAAI}, 2024.

\bibitem[Zhou et~al.(2017)Zhou, Zhao, Puig, Fidler, Barriuso, and Torralba]{zhouSceneParsingADE20K2017}
Bolei Zhou, Hang Zhao, Xavier Puig, Sanja Fidler, Adela Barriuso, and Antonio Torralba.
\newblock Scene {{Parsing Through ADE20K Dataset}}.
\newblock In \emph{CVPR}, 2017.

\bibitem[Zhou et~al.(2023)Zhou, Wang, Ye, and Zhan]{zhouModel603Exemplars2023}
Da-Wei Zhou, Qi-Wei Wang, Han-Jia Ye, and De-Chuan Zhan.
\newblock A {{Model}} or 603 {{Exemplars}}: {{Towards Memory-Efficient Class-Incremental Learning}}.
\newblock In \emph{ICLR}, 2023.

\bibitem[Zhu et~al.(2023)Zhu, Chen, Yin, See, and Liu]{zhuContinualSemanticSegmentation2023}
Lanyun Zhu, Tianrun Chen, Jianxiong Yin, Simon See, and Jun Liu.
\newblock Continual {{Semantic Segmentation}} with {{Automatic Memory Sample Selection}}.
\newblock In \emph{CVPR}, 2023.

\bibitem[Zhu et~al.(2021)Zhu, Su, Lu, Li, Wang, and Dai]{zhuDeformableDETRDeformable2021}
Xizhou Zhu, Weijie Su, Lewei Lu, Bin Li, Xiaogang Wang, and Jifeng Dai.
\newblock Deformable {{DETR}}: {{Deformable Transformers}} for {{End-to-End Object Detection}}.
\newblock In \emph{ICLR}, 2021.

\end{thebibliography}
}

\clearpage
\setcounter{page}{1}
\setcounter{table}{0}
\setcounter{figure}{0}
\setcounter{algorithm}{0}

\renewcommand{\thetable}{A\arabic{table}}
\renewcommand{\thefigure}{A\arabic{figure}}
\renewcommand{\thealgorithm}{A\arabic{algorithm}}
\maketitlesupplementary
\appendix


This supplementary material first presents a detailed workflow of the class recognition process. Next, it provides more comparisons with state-of-the-art methods. Finally, more ablation studies and detailed results are reported.

\section{Workflow of Stage 2 Class Recognition}

\begin{figure}[h]
    \centering
    \includegraphics[width=\columnwidth]{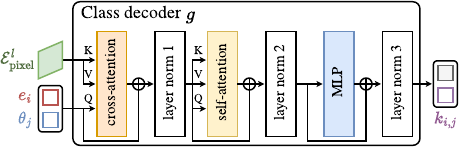}
    \caption{Class decoder architecture.}
    \label{fig:cls-decode}
\end{figure}

In detail, we present the class decoder architecture in~\cref{fig:cls-decode}. It is a single Transformer decoder block whose key (K) and value (V) are pixel embedding; query (Q) is positional embedding with task query. Note that only one positional embedding (red), together with one task query (blue), is fed through the class decoder once a time. The task embedding (purple) is the corresponding output of the task query, and the other embedding (gray) is discarded.

We also illustrate the stage 2 class recognition process in~\cref{algo:stage-2}. The class decoding process requires positional embeddings, task queries, and pixel embeddings as input and outputs the class probability of all proposals.

\begin{algorithm}
    \caption{Class recognition at step $t$}
    \label{algo:stage-2}
    \begin{algorithmic}[1]
        \Require{$\mathcal{E}^\mathrm{m}_\mathrm{pos} = \{e_1, \dots, e_M\}$: matched positional embeds\\
        $\mathcal{Q}_\mathrm{task}^t = \{\theta_1, \dots, \theta_t\}$: task queries at step $t$\\
        $\Phi^t = \{\phi_1, \dots, \phi_t\}$: classifiers at step $t$\\
        $\mathcal{E}^l_\mathrm{pixel}$: corresponding pixel embeddings\\
        $g$: class decoder}
        \Ensure{$\mathcal{P}^t$: class probability at step $t$}
        \For{$i \gets 1, \dots, M$}
            \For{$j \gets 1, \dots, t$}
                \State $\mathcal{Q}_\mathrm{cls}^{i, j} \gets (e_i, \theta_j)$ 
                \State $k_{i, j} \gets g(\mathcal{Q}_\mathrm{cls}^{i, j}, \mathcal{E}^l_\mathrm{pixel})$
                \State $z_{i, j} \gets \phi_j(k_{i, j})$
            \EndFor
            \State $z_i \gets [z_{i, 1}, \dots, z_{i, t}]$
            \State $p_i \gets \mathrm{sigmoid}(z_i)$
        \EndFor
        \State $\mathcal{P}^t \gets \{p_1, \dots, p_M\}$
    \end{algorithmic}
\end{algorithm}

\section{More Comparisons with State-of-the-arts}

We compare CoMasTRe with state-of-the-art continual segmentation methods. Here, we additionally report quantitative results in ADE20K \textit{50-50} setting and qualitative results of PASCAL VOC \textit{15-1} and ADE20K \textit{100-10}. 

\noindent\textbf{Quantitative results in ADE20K \textit{50-50} setting.}
As shown in~\cref{tab:ade-50-50}, we report benchmark results in ADE20K~\cite{zhouSceneParsingADE20K2017}~\textit{50-50} setting. The results show that we achieve a new state-of-the-art in this setting with 0.32 percent point (\textit{p.p}) leap on \textit{all} metric compared with CoMFormer~\cite{cermelliCoMFormerContinualLearning2023}. Specifically, our method reaches a competitive new class performance while maintaining better old knowledge (0.63 \textit{p.p} gain compared with CoMFormer). Furthermore, CoMasTRe performs better across all learning steps, surpassing CoMFormer by 3.12 \textit{p.p} and even can beat the highest per-pixel method PLOP~\cite{douillardPLOPLearningForgetting2021}~by 0.30 \textit{p.p} on \textit{avg}. 

\begin{table}
    \centering
    \caption[ade-50-50]{Benchmark results in ADE20K \textit{50-50} setting. The 1\textsuperscript{st} and 2\textsuperscript{nd} highest results are marked in \textbf{bold} and \underline{underline}.}
    \resizebox{\columnwidth}{!}{
        \begin{tabular}{ll|cccc}
        \toprule
        \multirow{2}{*}{Paradigm}  & \multirow{2}{*}{Method}                                 & \multicolumn{4}{c}{\textbf{50-50} (3 tasks)}                                                              \\
                                   &                                                         & \textit{1-50} &          \textit{51-150}              & \textit{all}            & \textit{avg}            \\ \midrule
        \multirow{6}{*}{Per-Pixel} & MiB~\cite{cermelliModelingBackgroundIncremental2020}    & 45.57                    & 21.01                      & 29.31                   & 38.98                   \\
                                   & SDR~\cite{michieliContinualSemanticSegmentation2021}    & 45.66                    & 18.76                      & 27.85                   & 34.25                   \\
                                   & PLOP~\cite{douillardPLOPLearningForgetting2021}         & 48.83                    & 20.99                      & 30.40                   & \underline{39.42}                   \\
                                   & REMINDER~\cite{phanClassSimilarityWeighted2022}         & 47.11                    & 20.35                      & 29.39                   & 39.26                   \\
                                   & RCIL~\cite{zhangRepresentationCompensationNetworks2022} & 48.30                    & 25.00                      & 32.50                   & ---                     \\
                                   & \textit{Joint}          & 51.21                    & 32.77                      & 39.00                   & ---                     \\ \midrule
        \multirow{4}{*}{Query}     & CoMFormer~\cite{cermelliCoMFormerContinualLearning2023} & \underline{49.20}                    & \textbf{26.60}                      & \underline{34.10}                   & 36.60                   \\
                                   & \textit{Joint}          & 53.40                    & 38.00                      & 43.10                   & ---                     \\ \cmidrule(l){2-6} 
                                   & \textbf{CoMasTRe} (ours)                                &  \textbf{49.83}                          & \underline{26.56}                            & \textbf{34.42}                         & \textbf{39.72}                        \\
                                   & \textit{Joint}          & 54.09                    & 39.49                      & 44.36                   & ---                     \\ \bottomrule
        \end{tabular}
    }
    \label{tab:ade-50-50}
\end{table}

\begin{figure*}
    \centering
    \includegraphics[width=\textwidth]{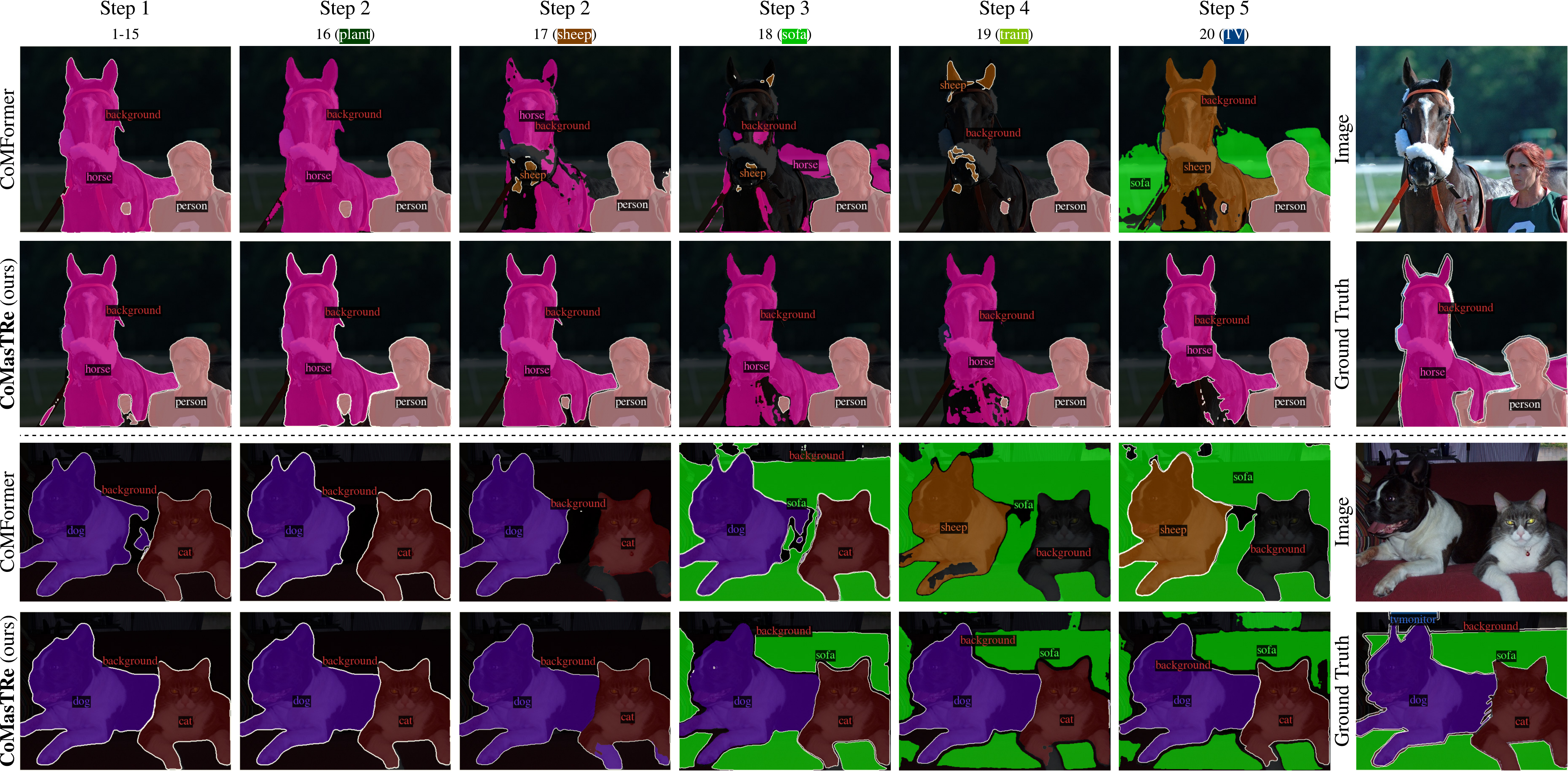}
    \caption{Qualitative results compared with CoMFormer~\cite{cermelliCoMFormerContinualLearning2023}~in PASCAL VOC \textit{15-1} setting.}
    \label{fig:pascal-vis}
\end{figure*}

\noindent\textbf{Qualitative results on PASCAL VOC 2012.}
Compared with CoMFormer~\cite{cermelliCoMFormerContinualLearning2023}, we visualize the segmentation results under PASCAL VOC \textit{15-1} in~\cref{fig:pascal-vis}. By comparison, our method is more resistant to forgetting old similar classes. For example, CoMFormer starts to forget the \textit{horse} after learning \textit{sheep} class at step 2 (first row) and misrecognizes the \textit{dog} as a \textit{sheep} at step 4 (third row). Additionally, our method is less prone to overconfidence on new classes, \eg, CoMFormer falsely recognizes a \textit{sofa} at step 5 (first row), but our method does not.

\begin{figure*}
    \centering
    \includegraphics[width=0.99\textwidth]{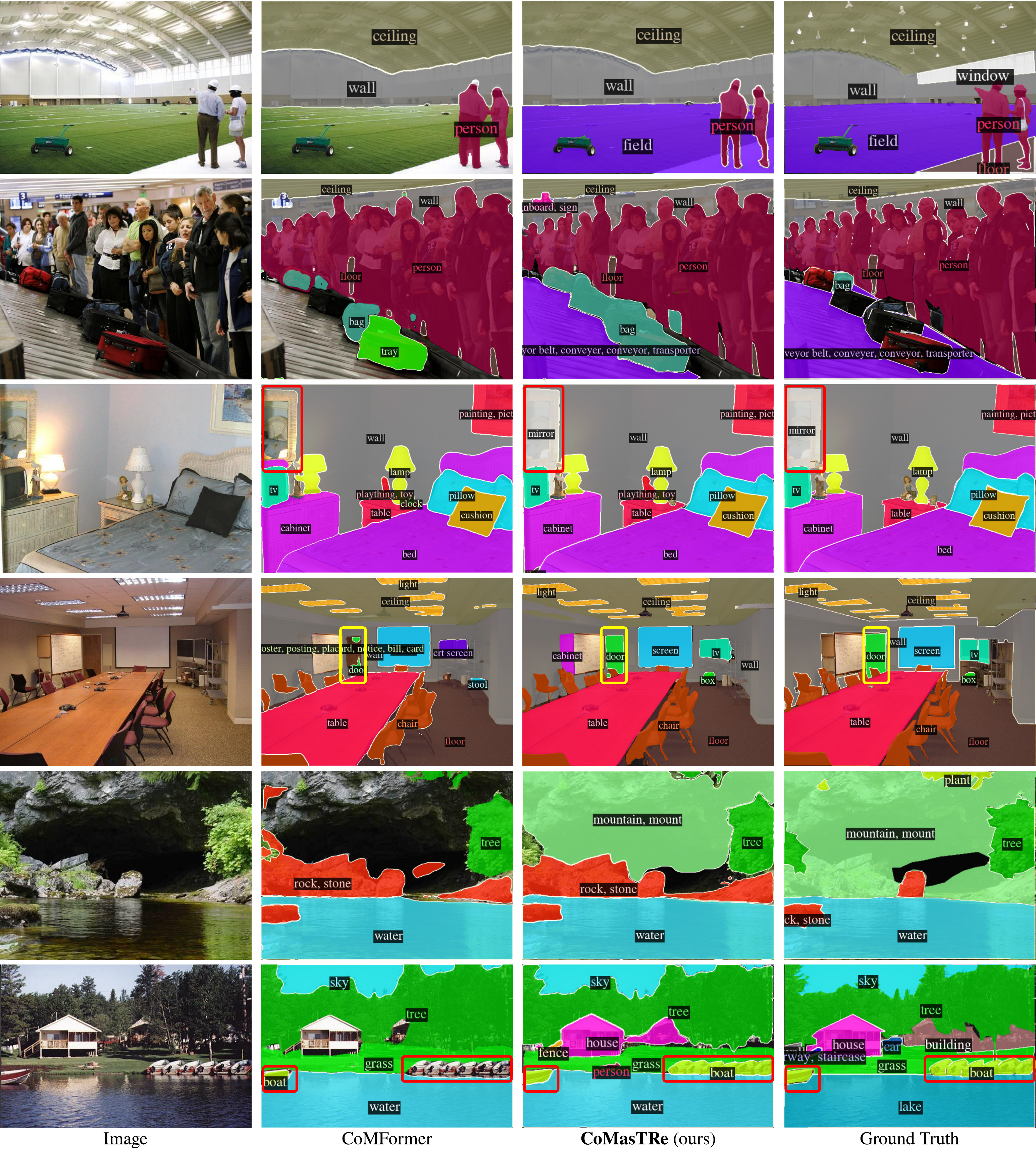}
    \caption{Qualitative results compared with CoMFormer~\cite{cermelliCoMFormerContinualLearning2023}~in ADE20K \textit{100-10} setting.}
    \label{fig:ade-vis}
\end{figure*}

\noindent\textbf{Qualitative results on ADE20K.}
As shown in~\cref{fig:ade-vis}, we visualize the results in ADE20K \textit{100-10} settings. The visualization suggests that our method preserves better knowledge of previous classes. Compared with CoMFormer~\cite{cermelliCoMFormerContinualLearning2023}, our method correctly segments the \textit{field} in the first row, the \textit{transporter} in the second row, the \textit{mirror} in the third row, the \textit{mountain} in the fifth row, and the \textit{house} in the sixth row. In addition, CoMasTRe also proposes better masks for old classes. For example, our method maintains the knowledge of the \textit{door} in the fourth row (yellow box) and the \textit{boat} in the sixth row (red boxes).

\section{More Ablation Studies and Detailed Results}

We present more ablation studies on the forgetting of objectness and the effectiveness of objectness score reweighting. In addition, per-class segmentation results on PASCAL VOC are reported.

\begin{table}
    \centering
    \caption[obj-forget]{Average recall (AR) of mask proposals in PASCAL VOC \textit{15-1} setting, where $\mathcal{L}_\mathrm{mask-kd}$ for mask distillation, $\mathcal{L}_\mathrm{os-kd}$ for objectness score distillation, $\mathcal{L}_\mathrm{pe-kd}$ for position distillation, $s$ for objectness score, and $\alpha$ for high objectness threshold.}
    \resizebox{\columnwidth}{!}{
        \begin{tabular}{c|l|cc}
            \toprule
            \multirow{2}{*}{\#} & \multirow{2}{*}{\textbf{Case}} & \multicolumn{2}{c}{\textbf{AR}} \\
                                           & & $s > 0$ & $s > \alpha$          \\ \midrule
            1 & $\mathcal{L}_\mathrm{mask-kd}$ & 55.89   & 13.75                 \\
            2 & $\mathcal{L}_\mathrm{os-kd}$   & 50.92   & 47.84                 \\
            3 & $\mathcal{L}_\mathrm{mask-kd} + \mathcal{L}_\mathrm{os-kd}$ & \underline{55.63}   & \underline{49.02}                 \\
            4 & $\mathcal{L}_\mathrm{mask-kd} + \mathcal{L}_\mathrm{os-kd} + \mathcal{L}_\mathrm{pe-kd}$ & \underline{55.84}   & \underline{48.83}                 \\ 
              \bottomrule
        \end{tabular}
    }
    \label{tab:obj-forget}
\end{table}

\noindent\textbf{Objectness forgetting analysis.}
We analyze the forgetting of objectness by ablating objectness distillation components. The ablation is conducted in the same cases as in Tab. \textcolor{red}{5} in the main text.
For each case, we use average recall (AR) to indicate the performance of mask proposals. Here, $s$ stands for the objectness score, and $\alpha$ is the objectness threshold during inference.
As shown in~\cref{tab:obj-forget}, we report the performance in PASCAL VOC~\cite{everinghamPascalVisualObject2010}~\textit{15-1} setting. By comparing case 1 and 3, without $\mathcal{L}_\mathrm{os-kd}$, AR ($s > 0$) remains unchanged, but AR ($s > \alpha$) diminishes (-35.27 \textit{p.p} AR), which means the objectness scores fail to indicate old class objectness.
By comparing case 2 and 3, we observe slight forgetting of mask proposals without $\mathcal{L}_\mathrm{mask-kd}$ (-4.71 \textit{p.p} AR), showing the forgetting robustness of mask proposals.
When comparing case 3 and 4, we find position distillation contributes most to continual classification (see Tab. \textcolor{red}{5} in the main text), as the AR changes are negligible (underlined in~\cref{tab:obj-forget}).

\begin{table}
    \centering
    \caption[obj-weight]{Ablation results of objectness score reweighting in PASCAL VOC \textit{15-1} setting by varying reweighting strength $\beta$. Note that no reweight is applied when $\beta = 0.0$.
    }
    \resizebox{0.8\columnwidth}{!}{
        \begin{tabular}{c|c|cccc}
        \toprule
        \# & $\beta$          & \textit{1-15} & \textit{16-20} & \textit{all} & \textit{avg} \\ \midrule
        1 & 0.0               & 67.47         & 41.83          & 61.37        & 68.47        \\
        2 & 1.0               & \textbf{69.79}  	     & 42.98          & 63.41        & 70.35        \\
        3 & 2.0               & 69.77         & \textbf{43.62}          & \textbf{63.54}        & \textbf{70.63}        \\
        \bottomrule
        \end{tabular}
    }
    \label{tab:obj-reweight}
    \vspace{-1.5mm}
\end{table}

\begin{table*}
    \centering
    \caption[per-class]{Per-class segmentation results on PASCAL VOC 2012 in mIoU (\%). Incremented classes (\textit{inc}) are marked in \textcolor{OliveGreen}{green}.}
    \resizebox{0.991\textwidth}{!}{
        \begin{tabular}{l|ccccccccccccccccccccc|ccc}
        \toprule
        Setting        & bg   & aero & bike & bird & boat & bottle & bus  & car  & cat  & chair & cow  & table & dog  & horse & mbike & person & plant & sheep & sofa & train & tv   & \textit{base} & \textit{inc}  & \textit{all}  \\
        \midrule
        \textbf{19-1} (2 steps) & 93.7 & 91.9 & 42.8 & 88.9 & 65.4 & 81.1   & 87.9 & 80.7 & 91.9 & 37.0  & 80.1 & 51.3  & 86.0 & 82.9  & 82.0  & 87.0   & 56.6  & 87.3  & 50.6 & 77.4  & \textcolor{OliveGreen}{69.5} & 75.1 & 69.5 & 74.9 \\
        \textbf{15-5} (2 steps) & 93.5 & 90.4 & 43.6 & 90.8 & 64.2 & 82.8   & 88.2 & 88.6 & 94.1 & 42.8  & 80.9 & 70.5  & 89.4 & 82.7  & 84.6  & 88.4   & \textcolor{OliveGreen}{39.4}  & \textcolor{OliveGreen}{59.1}  & \textcolor{OliveGreen}{38.9} & \textcolor{OliveGreen}{62.0}  & \textcolor{OliveGreen}{60.2} & 79.7 & 51.9 & 73.1 \\
        \textbf{15-1} (6 steps) & 88.9 & 86.4 & 38.0 & 82.6 & 53.2 & 76.8   & 76.8 & 83.2 & 82.5 & 36.2  & 59.3 & 48.4  & 80.4 & 63.1  & 74.7  & 85.8   & \textcolor{OliveGreen}{29.9}  & \textcolor{OliveGreen}{47.1}  & \textcolor{OliveGreen}{33.0} & \textcolor{OliveGreen}{55.4}  & \textcolor{OliveGreen}{52.7} & 69.8 & 43.6 & 63.5 \\
        \midrule
        \textit{Joint} & 94.3 & 91.8 & 43.0 & 91.1 & 65.3 & 85.2   & 92.2 & 87.2 & 93.1 & 44.2  & 85.3 & 69.4  & 90.5 & 86.9  & 85.3  & 88.4   & 60.7  & 83.9  & 48.1 & 85.4  & 68.3 & ---  & ---  & 78.1 \\
        \bottomrule
        \end{tabular}
    }
    \label{tab:per-class}
\end{table*}

\noindent\textbf{Effectiveness of objectness score reweighting.}
In~\cref{tab:obj-reweight}, we ablate the effectiveness of objectness score reweighting mentioned in Sec. \textcolor{red}{3.3.1} of the main text. The reweight strength $\beta$ is set to 0.0, 1.0, and 2.0, respectively. Please note that when $\beta$ is set to 0.0, the distillation is equivalent to the regular unweighted one.
The results show that the reweighting ($\beta = 2.0$) leads to 2.17 \textit{p.p} performance gain on \textit{all} metric compared with the unweighted version.

\noindent\textbf{Per-class results on PASCAL VOC.}
In \cref{tab:per-class}, we provide per-class experimental results on PASCAL VOC 2012 in different continual segmentation settings. The results indicate that in addition to learning new classes, the old class performance can be further strengthened in later learning steps compared with \textit{joint} baseline, such as \textit{sheep} class in \textit{19-1} (+3.4 \textit{p.p}) and \textit{car} class in \textit{15-5} (+1.4 \textit{p.p}).

\end{document}